\documentclass[journal,twoside,web]{ieeecolor}
\usepackage{generic}
\usepackage{graphicx}
\usepackage{cite}
\usepackage{amsmath,amssymb,amsfonts}
\usepackage{multirow}
\usepackage{verbatim}
\usepackage{algorithm}
\usepackage{algpseudocode}
\usepackage{array}
\usepackage{textcomp}

\DeclareMathOperator*{\argmax}{arg\,max}

\begin{document}


\twocolumn
\pagenumbering{arabic}
\setcounter{page}{1}

\def\BibTeX{{\rm B\kern-.05em{\sc i\kern-.025em b}\kern-.08em
    T\kern-.1667em\lower.7ex\hbox{E}\kern-.125emX}}
\markboth{\journalname, VOL. XX, NO. XX, XXXX 2019}
{Qin \MakeLowercase{\textit{et al.}}: Varifocal-Net: A Chromosome Classification Approach using Deep Convolutional Networks (March 2019)}

\title{Varifocal-Net: A Chromosome Classification Approach Using Deep Convolutional Networks}
\author{Yulei Qin, \IEEEmembership{Student Member, IEEE}, Juan Wen, Hao Zheng, Xiaolin Huang, \IEEEmembership{Senior Member, IEEE}, Jie Yang, Ning Song, Yue-Min Zhu, Lingqian Wu, Guang-Zhong Yang, \IEEEmembership{Fellow, IEEE}
\thanks{The first two authors contributed equally to this work. \emph{Corresponding author: Jie Yang, Lingqian Wu, and Ning Song.} This work was partially supported by National Natural Science Foundation of China (NSFC, 61603248, 61572315, 6151101179), 863 Plan of China 2015AA042308, 973 Plan of China 2015CB856004, 1000-Talent Plan (Young Program) and Committee of Science and Technology, Shanghai, China (No. 17JC1403000). Copyright (c) 2019 IEEE. Personal use of this material is permitted. However, permission to use this material for any other purposes must be obtained from the IEEE by sending a request to pubs-permissions@ieee.org.}
\thanks{Y. Qin, H. Zheng, X. Huang, and J. Yang are with the Institute of Image Processing and Pattern Recognition, Shanghai Jiao Tong University, Shanghai 200240, China (e-mail: jieyang@sjtu.edu.cn).}
\thanks{J. Wen and L. Wu are with the Center for Medical Genetics, School of Life Sciences, Central South University, Changsha 410078, China (e-mail: wulingqian@sklmg.edu.cn).}
\thanks{N. Song is with the Shanghai Key Laboratory of Reproductive Medicine, School of Medicine, Shanghai Jiao Tong University, Shanghai 200025, China and also with the Diagens-Hangzhou, Hangzhou 311121, China (e-mail: ningsong@shsmu.edu.cn).}
\thanks{Y.-M. Zhu is with the University Lyon, INSA Lyon, CNRS, INSERM, CREATIS UMR 5220, U1206, F-69621, France.}
\thanks{G.-Z. Yang is with the Hamlyn Centre for Robotic Surgery, Imperial College London, SW72AZ, UK.}
}

\maketitle

\begin{abstract}
Chromosome classification is critical for karyotyping in abnormality diagnosis. {To expedite the diagnosis}, we present a novel method named Varifocal-Net for simultaneous classification of chromosome’s type and polarity using deep convolutional networks. The approach consists of one global-scale network (G-Net) and one local-scale network (L-Net). It follows {three} stages. The first stage is to learn both global and local features. We extract global features and detect finer local regions via the G-Net. {By proposing a} varifocal mechanism, we zoom into local parts and extract local features via the L-Net. Residual learning and multi-task learning strategies are utilized to promote high-level feature extraction. The detection of discriminative local parts is fulfilled by a localization subnet of the G-Net, whose training process involves both supervised and weakly-supervised learning. The second stage is to build two multi-layer perceptron classifiers that exploit features of both two scales to boost classification performance. {The third stage is to introduce a dispatch strategy of assigning each chromosome to a type within each patient case, by utilizing the domain knowledge of karyotyping.} Evaluation results from 1909 karyotyping cases {showed that the proposed Varifocal-Net achieved the highest accuracy per patient case (\%) of 99.2 for both} type and polarity tasks. It outperformed state-of-the-art methods, demonstrating the effectiveness of our varifocal mechanism, multi-scale feature ensemble, and dispatch strategy. {The proposed method has been applied to assist practical karyotype diagnosis.}
\end{abstract}


\section{Introduction}
\label{sec:introduction}
\begin{figure}[htbp]
\centerline{\includegraphics[width=\columnwidth]{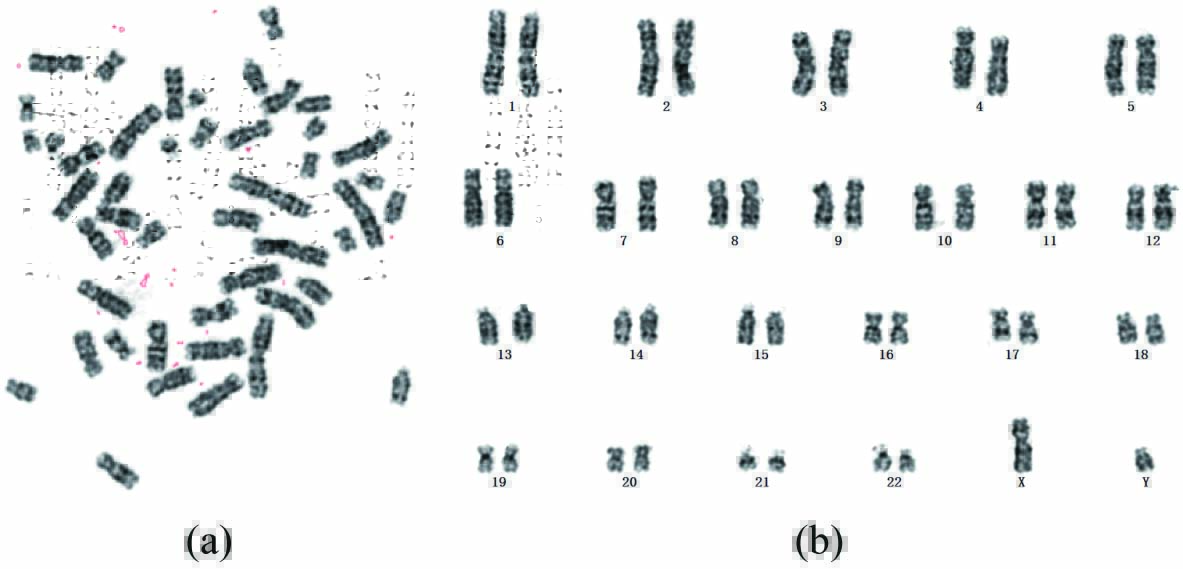}}
\caption{(a) A Giemsa-stained microscopic image of male chromosomes for one case. (b) The karyotyping result {map} of (a) is formed of the paired and ordered chromosomes (22 pairs of autosomes and 1 pair of sex chromosomes XY).}
\label{karyotype}
\end{figure}

\IEEEPARstart{C}{hromosome} anomalies, including numerical and structural abnormalities, are responsible for several genetic diseases such as leukemia \cite{natarajan2002chromosome}. Numerical abnormalities arise from the gain or loss of an entire chromosome, which constitute a great proportion of abnormalities \cite{theisen2010disorders}. Structural abnormalities result from the breakage and reunion of chromosome segments. In clinical practice, an important procedure for chromosome diagnosis is karyotyping, which is carried out on microscopic images of a single cell \cite{piper1990automated}. {The karyotyping requires first using staining techniques on each cell to obtain stained meta-phase chromosomes. These chromosomes are classified by operators and sorted into 22 pairs of autosomes and 1 pair of sex chromosomes (XX or XY) in the karyotyping map. Then, doctors analyze such map for diagnosis. The karyotyping can be classified into two main categories by the used staining technique: Giemsa karyotyping using Giemsa staining and fluorescent karyotyping using fluorescent staining. If fluorescent staining is employed together with deconvolution of fluorescence signals, chromosomes of different types will be dyed with different colors for fluorescent karyotyping (e.g., SKY \cite{schrock1996multicolor} and M-FISH \cite{speicher1996karyotyping}). If Giemsa staining is adopted, banding patterns that appear alternatively darker and lighter gray-levels (e.g., G-bands) will be produced for Giemsa karyotyping.} {Although fluorescent karyotyping is easy for operators to distinguish chromosomes by color, its inherent limitations (e.g., difficulty of detecting all chromosomal abnormalities, impermanent preservation of fluorescence signals, prohibitive cost, controversial reliability of probe hybridization, and unavailability of various probes and clinical samples) make it inappropriate as a first-tier screening tool for examinations \cite{lee2001limitations,huber2018fluorescence,gozzetti2000fluorescence}. In contrast, Giemsa karyotyping can detect nearly all abnormalities with a single low-cost test, making it preferred in practice compared to fluorescent karyotyping.} A typical karyotyping result map from Giemsa-stained chromosomes is shown in Fig. \ref{karyotype}.

The process of karyotyping demands meticulous efforts from well-trained operators. To reduce the burden of karyotyping, many automated classification methods have been developed for analyzing meta-phase chromosomes \cite{lerner1995medial, ming2010automatic, markou2012automatic, madian2014analysis, 1438347, abid2018survey, sharma2017crowdsourcing, gupta2017siamese, wu2018end}. In general, such methods consist of three steps. The first is to preprocess the chromosome image, which usually involves skeletonization algorithms to compute the medial axis of each chromosome in the image. The second is to extract features along each computed axis. The third step is to build classifiers (e.g., multi-layer perceptron (MLP) and support vector machine (SVM)) to estimate chromosome's type based on the extracted features.

Traditional classification methods mainly rely on geometrical features (e.g., {a chromosome's length, centromere position, and banding pattern features).} Lerner \textit{et al.} \cite{lerner1995medial} first proposed two approaches of computing medial axis transform (MAT) to detect medial axes of chromosomes. Then, intensity-based features and centromeric indexes were fed into an MLP network for classification. Ming \textit{et al.}\cite{ming2010automatic} computed medial axes using a middle point algorithm. They extracted banding patterns by average intensity, gradient, and shape profiles and adopted an MLP classifier. Markou \textit{et al.} \cite{markou2012automatic} proposed a robust method to first extract medial axes using a thinning algorithm. Bifurcations of the axis were removed iteratively via a pixel-neighborhood-based pruning algorithm. Then, the axis was smoothed and extended, with the band-profile features extracted along it. An SVM classifier was finally adopted for type classification. Several other methods targeted at precise detection of the medial axis and centromere location \cite{stanley1996centromere, wang2008rule, arachchige2013intensity, loganathan2013analysis}, providing a foundation for accurate chromosome classification.

With the advent of deep learning, researchers tended to employ convolutional neural networks (CNNs) for feature extraction in {classification tasks \cite{lecun1998gradient, krizhevsky2012imagenet, szegedy2016rethinking,  simonyan2014very, he2016deep, huang2017densely, lin2015bilinear, fu2017look, jaderberg2015spatial}}. Three methods were reported on using deep learning techniques in chromosome studies. Sharma \textit{et al.} \cite{sharma2017crowdsourcing} proposed a CNN-based method for classification. Bent chromosomes were first straightened by cropping and stitching, and then normalized by length. The accuracy of classification was 86.7$\%$ for such preprocessed chromosomes. Gupta \textit{et al.} \cite{gupta2017siamese} developed a classification method based on the Siamese Network. Chromosomes were first straightened using two proposed approaches and then fed into the Siamese Network for high-level feature embeddings. An MLP classifier exploited such embeddings for classification and an average accuracy of 85.6$\%$ was achieved. Very recently, Wu \textit{et al.} \cite{wu2018end} proposed a VGG-Net-D-based approach for category classification. Due to inadequate labeled data, they adopted generative adversarial network (GAN) to generate samples as data augmentation. Their performance was far below requirement for clinical application, {with an average} precision of 63.5$\%$ achieved.

Although many microscopes are nowadays equipped with chromosome classification systems (e.g., {CytoVision \cite{micci2001complete,yang2010fish,rodahl2005chromosomal}, Ikaros \cite{gadhia2014rare}, and ASI HiBand \cite{fan2000sensitivity}}), users still have to manually drag each chromosome image and drop it to the target position in the practical karyotyping process due to their poor performance. Research studies reveal that the challenges of chromosome classification mainly lie in the following aspects: 1) Chromosomes are often curved and bent due to their non-rigid nature, making it difficult to accurately extract their medial axes. Hence, errors accumulate in the process of straightening and feature computation along such axes, leading to an accuracy drop. 2) Even for the chromosomes of the same class, they vary slightly from person to person in terms of local details. The generalizability and performance of traditional methods, which depend on manually designed features, may degrade for clinical applications. 3) The chromosome polarity, which reflects whether a chromosome's q-arm (long arm) is downward or upward, is often not considered in previous work. However, it is important to decide the chromosome's orientation in the process of repositioning for the karyotyping {map generation}. All the q-arms should stay downward.

\begin{figure}[htbp]
\centerline{\includegraphics[scale=0.35]{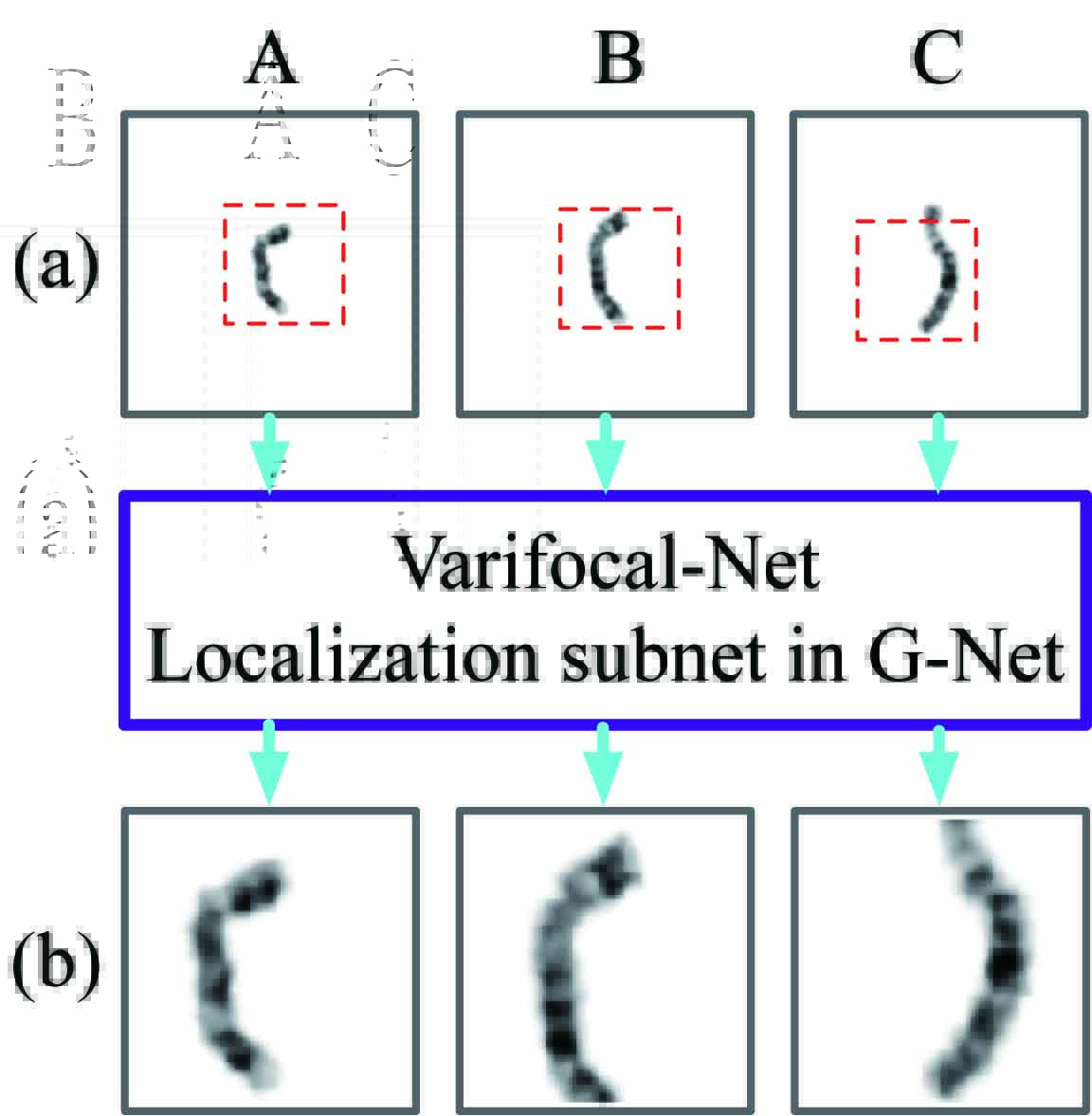}}
\caption{The focus is varied from global to local. Given chromosome images (A, B, C), the localization subnet detects their finer regions to crop and magnify. (a) The original chromosome images. (b) The local parts after zooming in.}
\label{glscale}
\end{figure}

To tackle the above challenges, we propose a novel CNN-based approach for chromosome classification. Its name, Varifocal-Net, highlights the capacity to zoom into local regions automatically. It has one global-scale network (G-Net) and one local-scale network (L-Net). We extract global features and pinpoint specific local regions via the G-Net. The view is changed (see Fig. \ref{glscale}) as our Varifocal-Net zooms into the discriminative region of a chromosome. Local features are extracted from such local parts via the L-Net. {At first glance, such a global-to-local idea resembles the concept of multi-scale CNNs used in cellular image analysis \cite{godinez2017multi, buyssens2012multiscale, godinez2018unsupervised, pan2018cell} and other vision tasks \cite{shen2015multi, zeng2017multi, lotter2017multi}. However, unlike previous multi-scale methods, our approach learns multi-scale information in the global-to-local mechanism. It locates the discriminative local region and extracts the features of the two scales through two independent networks.} The proposed {Varifocal-Net} comprises {three} stages. The first stage is to learn effective feature representations at both global and local scales. The global-scale representations mainly concern overall information such as the chromosome's length, shape, and size, which determines its type on a coarse-grained level. The local-scale representations depict details such as texture patterns of local parts, which facilitate discrimination among chromosomes on a fine-grained level. The second stage is to build two MLP classifiers to leverage features of both two scales for prediction of type and polarity, respectively. {The third stage is to introduce a dispatch strategy for type assignment within each patient case.} To validate the effectiveness and generalizability of our approach, we construct a large dataset containing 1909 karyotyping cases. Extensive experiments on the dataset corroborate that the Varifocal-Net {achieved} better performance than state-of-the-art methods. Our contributions can be summarized as follows:
\begin{itemize}
\item Inspired by the zoom capability of cameras, we propose the Varifocal-Net to address the challenges of chromosome classification. We extract global-scale features from the whole image and local-scale features from the local region selected by our varifocal mechanism. Residual learning and multi-task learning strategies are utilized to promote effective feature learning. The detection of discriminative local parts is fulfilled via a localization subnet whose training involves both supervised and weakly-supervised learning.
\item We utilize the concatenated features from both global and local scales to predict type and polarity simultaneously, thereby combining the knowledge acquired at two scales. To our best knowledge, this represents the first attempt to take multi-scale feature ensemble into account in chromosome studies.
\item {We propose a dispatch strategy to assign each chromosome to a type based on its predicted probabilities. Both the maximum likelihood criterion and possible abnormality situations are taken into account to enable the strategy suitable for clinical settings.}
\item We evaluate the proposed {approach} on a large dataset. It demonstrates its superior performance compared with state-of-the-art methods. The end-to-end manner of classification sidesteps the problem of inaccurate medial axis extraction and chromosome straightening.
\item The Varifocal-Net has been put into clinical practice for chromosome classification. For each patient, it accurately classifies both abnormal and healthy chromosomes and diagnoses numerical abnormalities if the number of classified chromosomes is irregular.
\end{itemize}

The paper is structured as follows: In Section \ref{sec:methods}, we describe the proposed method. In Section \ref{sec:experiment}, we provide experiments and results. Section \ref{sec:discussion} discusses our findings, followed by the conclusion in Section \ref{sec:conclusion}.

\section{Methods}
\label{sec:methods}
\begin{figure}[htbp]
\centerline{\includegraphics[width=\columnwidth]{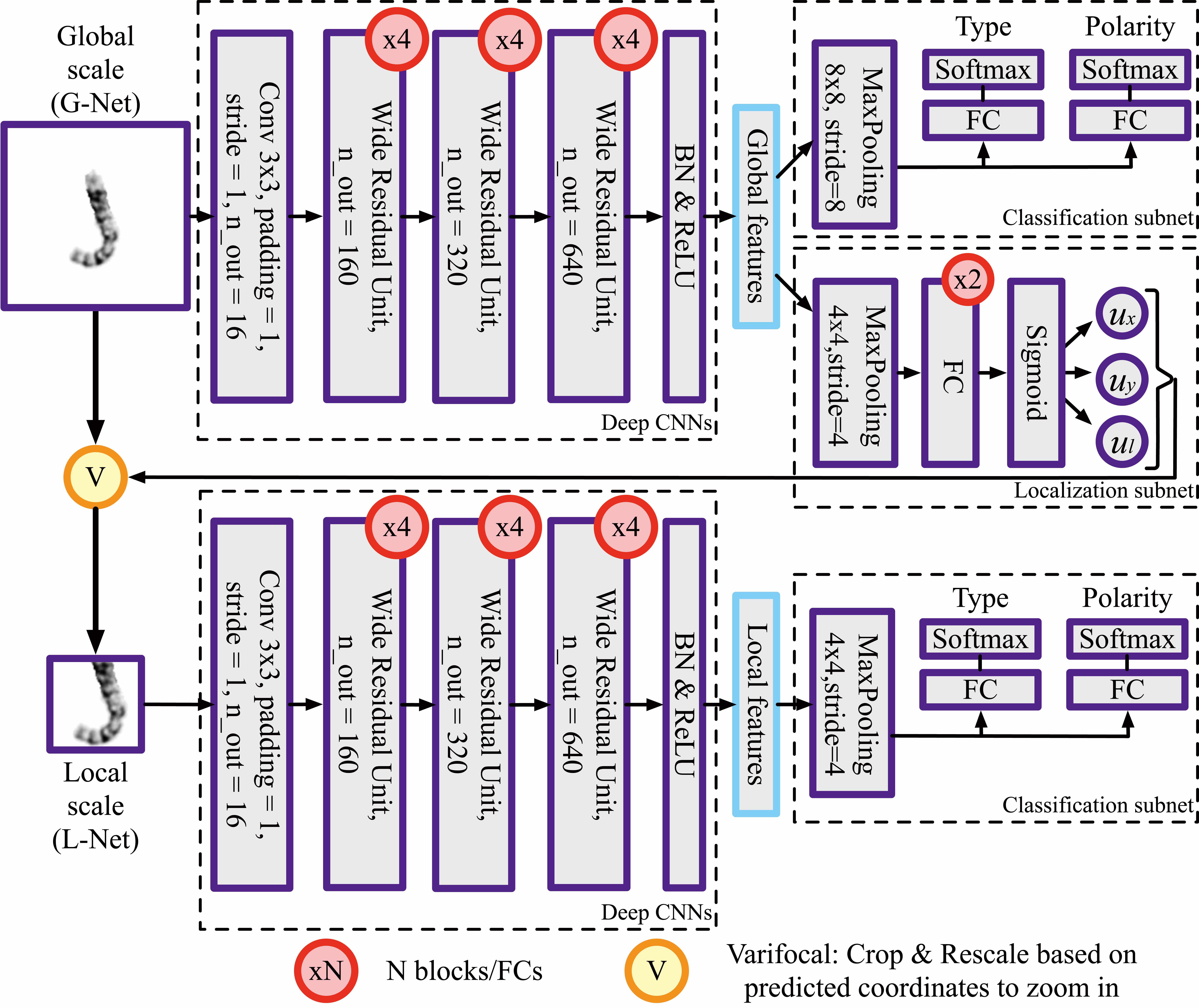}}
\caption{The first stage of the proposed Varifocal-Net: global-scale and local-scale feature extraction via the G-Net and the L-Net, respectively.
}
\label{varifocal-net}
\end{figure}

The proposed Varifocal-Net is composed of {three} stages: a) Global-scale and local-scale feature learning by optimizing the Varifocal-Net in an alternative way; b) Classification of type and polarity via MLP classifiers utilizing the fused features; {c) Assignment of chromosomes' types with the proposed dispatch strategy.} {Original chromosome images are separated manually by cytogeneticists from captured microscopic images. They are preprocessed as discussed in Sec. \ref{sec:imdetail} and taken as inputs to the G-Net in the first stage.} {The G-Net contains deep CNNs, one classification subnet, and one localization subnet, as shown in Fig. \ref{varifocal-net}. Global-scale features are extracted via the CNNs, which are optimized by the loss function of the classification subnet. After the CNNs and classification subnet converge, we pre-train the} localization subnet to output initial coordinates for local region detection. Then, with local parts cropped and rescaled, we optimize the L-Net and the localization subnet of the G-Net alternatively. In the second stage, with the fused two-scale features, we build two MLP classifiers to predict chromosome's type and polarity, respectively. The schematic representations of the first stage and the second stage of our Varifocal-Net are illustrated in Fig. \ref{varifocal-net} and Fig. \ref{totalpredict}, respectively. {For each chromosome within one patient case, a dispatch strategy is employed in the third stage to assign it to a certain type based on its predicted probabilities.}

\subsection{Stage 1: Global-scale and Local-scale Feature Learning}

\subsubsection{Feature Extraction with Residual Learning}
\begin{figure}[htbp]
\centerline{\includegraphics[scale=0.4]{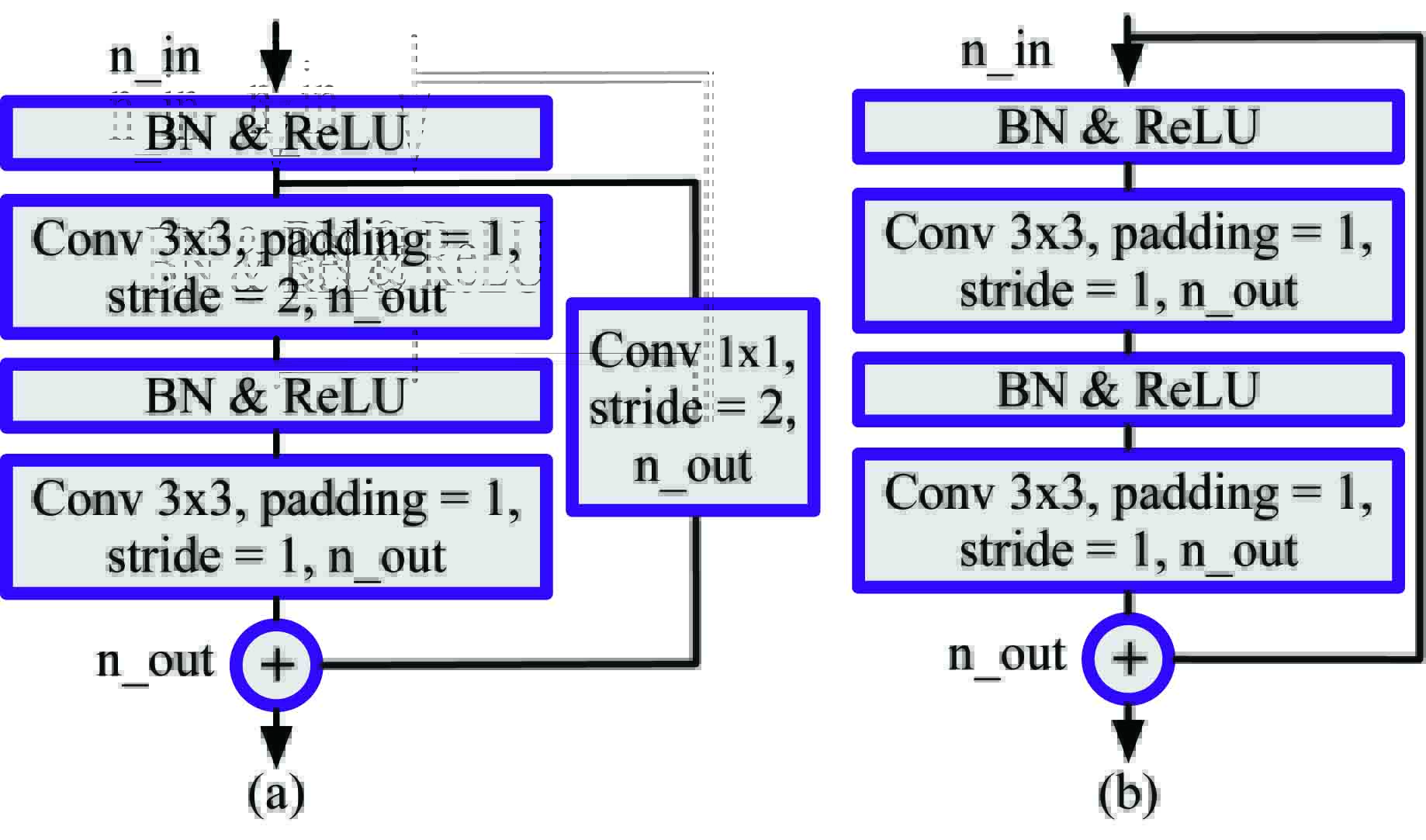}}
\caption{Wide residual unit. n$\_$in and n$\_$out stand for number of input and output feature channels, respectively. (a) if n$\_$in $\neq$ n\_out. (b) if n$\_$in $=$ n$\_$out.
}
\label{wideresblock}
\end{figure}

The architecture of deep CNNs for feature extraction is the same for both the G-Net and the L-Net. Inspired by the success of ResNet \cite{he2016deep, zagoruyko2016wide}, we adopt wide residual blocks to introduce residual learning. Such CNNs consist of one convolution layer (Conv), three residual blocks, one batch normalization layer (BN), and one rectified linear unit (ReLU). Each residual block has four residual units as illustrated in Fig. \ref{wideresblock}, with the first unit increasing {the number of channels} and downsampling features through strided convolution.

\subsubsection{Multi-task Learning with Weighted Classification Loss}
Since the tasks of type and polarity classification are correlated, we adopt multi-task learning to take inner relation between these tasks into consideration. It improves the effectiveness of feature extraction through a shared representation of CNNs \cite{caruana1997multitask}. In the classification subnet, a max-pooling layer is followed by two fully-connected (FC) layers respectively to predict type and polarity. The FC layers map the feature vector to the probability vectors of $24$ dimensions (for the type task) and $2$ dimensions (for the polarity task). We train the deep CNNs in the G-Net and the L-Net independently by minimizing a weighted loss of the classification subnet. For the type task, given a set of $N$ training {triplets} $\{(x_{i}, y_{i}^{t}, y_{i}^{p})\}_{i=1,2,...,N}$, the {cross-entropy} loss between the output vector $\mathbf{O^t}$ and the target vector $\mathbf{Y^t}$ is given by:
\begin{equation}
\mathcal{L}_{t}(\mathbf{O^t, Y^t}) = \sum_{i=1}^{N}-log(\frac{exp(o_i^t[y_i^t])}{\sum_{j=1}^{24}exp(o_i^t[j])}),
\label{eq:losstype}
\end{equation}
\noindent where $o_i^t$ and $y_i^t$ denote the output probability vector and the target type for the sample $x_i$, respectively. {Note that here we combine the softmax function and the standard cross-entropy function into one formula.} Similarly, the polarity classification loss between the predicted vector $\mathbf{O^p}$ and the target vector $\mathbf{Y^p}$ is defined as:
\begin{equation}
\mathcal{L}_{p}(\mathbf{O^p, Y^p}) = \sum_{i=1}^{N}-log(\frac{exp(o_i^p[y_i^p])}{\sum_{j=1}^{2}exp(o_i^p[j])}),
\label{eq:losspolarity}
\end{equation}
\noindent where $o_i^p$ and $y_i^p$ stand for the probability vector and the target polarity, respectively. The total multi-task loss is given by:
\begin{equation}
\mathcal{L}_{cls}(\mathbf{O^t, Y^t, O^p, Y^p}) = \mathcal{L}_{t}(\mathbf{O^t, Y^t}) + \lambda\mathcal{L}_{p}(\mathbf{O^p, Y^p}),
\label{eq:weightedloss}
\end{equation}
\noindent in which $\lambda$ is a weight controlling the balance between the two loss terms. We place more emphasis on the type task, thus setting $\lambda = 0.5$ in our experiments.

\subsubsection{Varifocal Mechanism}
\begin{figure}[htbp]
\centerline{\includegraphics[scale=0.15]{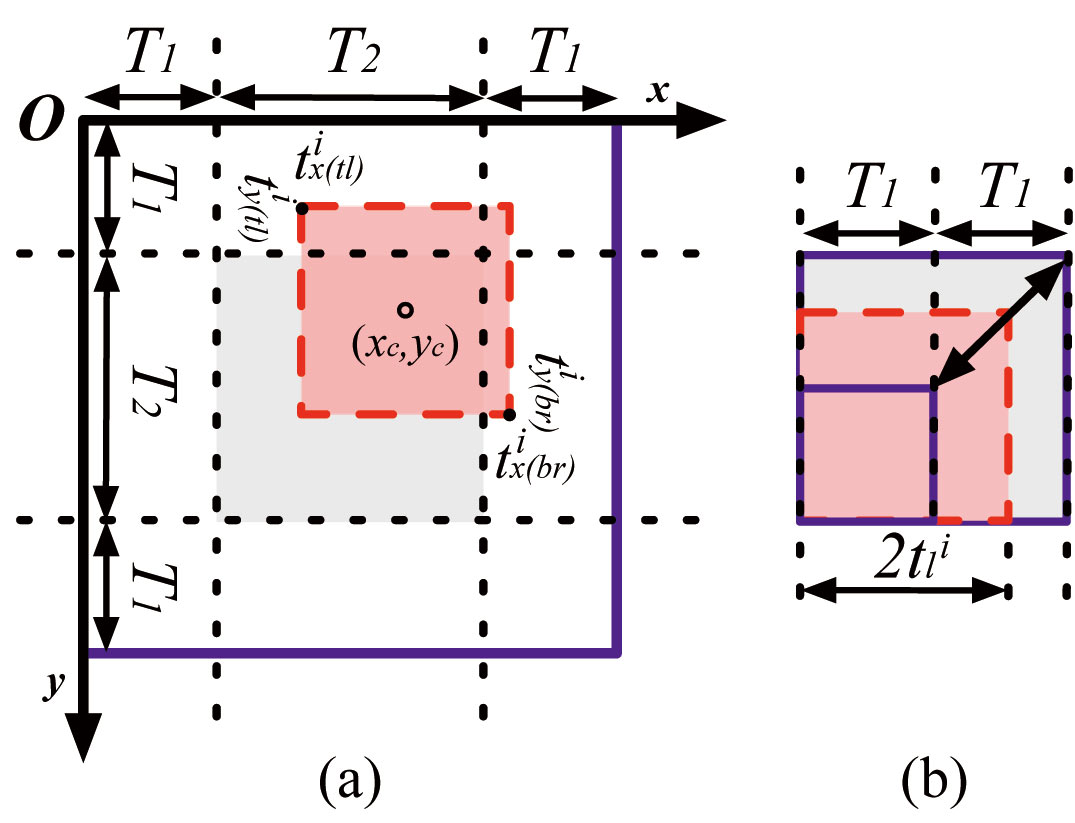}}
\caption{The diagram of parameterizations for the sample $x_i$. (a) The red box is the predicted local region and the gray background square is the area where the box's center pixel $(x_c,y_c)$ can {be located}. (b) The side length of the predicted box $(2t_{l}^i)$ is restricted, ranging from $T_1$ to $2T_1$.}
\label{fig:pos}
\end{figure}

Previous work on chromosome classification takes no advantage of multi-scale feature learning and fusing. These methods do not detect specific finer parts for detail description (e.g., nuance of banding's number, width, and intensity among similar chromosomes). Motivated by the success of region proposal network (RPN) \cite{girshick2015fast, ren2015faster} and attention proposal network (APN) \cite{fu2017look}, we propose a varifocal mechanism that zooms into local regions of chromosomes automatically for finer feature extraction. Given a chromosome sample $x_i$, it first predicts the position and size of a local region box via the localization subnet, which is sequentially composed of a max-pooling layer, two FC layers, and a sigmoid layer. The square box prediction is {expressed} as:
\begin{equation}
(u_x^i, u_y^i, u_l^i) = f(\mathbf{W_c}*x_i),
\label{eq:boxpredict}
\end{equation}
\noindent where $\mathbf{W_c}$ and $*$ denote all parameters of deep CNNs and their related operations (e.g., Conv, BN, and ReLU), respectively. $\mathbf{W_c}*x_i$ gives the global feature of $x_i$ and $f(\cdot)$ represents the proposed localization subnet. The variables $u_x^i$ and $u_y^i$ denote the relative coordinates of the box's center $(x_c, y_c)$ and $u_l^i$ is the relative length of the half of its side. All these variables range from $0$ to $1$. Assuming the top-left corner of $x_i$ as the origin of the global pixel coordinate system where $x$-axis starts from left to right and $y$-axis from top to bottom, we adopt the parameterizations of the top-left ($tl$) and bottom-right ($br$) pixels of the region box as follows:
\begin{equation}
\begin{aligned}
t_{x(tl)}^i = T_{1} + u_x^i\cdot T_{2} - t_{l}^i,\ t_{x(br)}^i &= T_{1} + u_x^i\cdot T_{2} + t_{l}^i,\\
t_{y(tl)}^i = T_{1} + u_y^i\cdot T_{2} - t_{l}^i,\ t_{y(br)}^i &= T_{1} + u_y^i\cdot T_{2} + t_{l}^i,\\
t_{l}^i &= u_l^i\cdot T_{1}/2 + T_{1}/2,
\end{aligned}
\label{eq:position}
\end{equation}
\noindent where $T_1, T_2$, and $t_{l}^i$ denote the minimum margin, maximum shift, and half of the side length, respectively. Fig. \ref{fig:pos} illustrates these parameterizations. Note that here we restrict the position and size of the predicted local region for two reasons. First, the predicted region should focus on a discriminative part of the chromosome, which is in the center of the image. Second, the region cannot {exceed the} image boundary and its size should be moderate to effectively capture local features. In our implementation, we set $T_2 = 2T_1$ empirically because it forces the localization subnet to focus on the central region.

Once a local region is predicted, the focus is moved onto it by cropping and rescaling. The cropping operation is implemented using a variant of {two-dimensional (2-D)} boxcar function \cite{fu2017look} as an approximation. Given the coordinate tuple $(t_{x(tl)}^i, t_{y(tl)}^i, t_{x(br)}^i, t_{y(br)}^i)$, we use the boxcar function to generate a region mask and multiply it with the original image in an element-wise manner. It is mathematically expressed as:
\begin{equation}
\begin{aligned}
x_{i}^{loc} &= x_i \odot boxcar(t_x^i, t_y^i, t_{l}^i),\\
boxcar(t_x^i, t_y^i, t_{l}^i) &= (H(x-t_{x(tl)}^i)-H(x-t_{x(br)}^i))\\
&\cdot(H(y-t_{y(tl)}^i)-H(y-t_{y(br)}^i))\\
\end{aligned}
\label{eq:crop}
\end{equation}
\noindent where $\odot$ denotes element-wise multiplication and $x_i^{loc}$ stands for the cropped local part. The 2-D $boxcar(t_x^i, t_y^i, t_{l}^i)$ function serves as a mask and $H(x)$ is the Heaviside step function. Note that the derivative of $H(x)$ is infinite at $x=0$. Since its derivative is required in back-propagation, we use the logistic function as a smooth analytic approximation for $H(x)$ in experiments, which is computed by:
\begin{equation}
\begin{aligned}
H(x) = \frac{1}{1+e^{-kx}},\ k>0
\end{aligned}
\label{eq:logistic}
\end{equation}
\noindent in which a larger $k$ (e.g., $k = 10$) leads to a sharper change at $x=0$. The multiplication with $boxcar(t_x^i, t_y^i, t_{l}^i)$ will mask out the target local region by keeping the value of pixels inside the region almost unchanged and that of others close to zero. Then, we {crop the target region in $x_{i}^{loc}$ and rescale it} to a unified size via bilinear interpolation, which makes it easier for both algorithm implementation and finer feature extraction in the L-Net. So far, the Varifocal-Net has zoomed into a particular local part. {Note that in the forward process, the local region is cropped directly by indexed slicing. In the backward propagation process, since the cropping operation is not derivative, the boxcar function is used to approximate it and provide necessary gradient for proper parameter optimization. Detailed analytical derivations are presented in Sec. \ref{sec:backprop}.}

\subsubsection{Loss Function of the Localization Subnet}
With definitions of the localization subnet $f(\cdot)$, we adopt both supervised and weakly-supervised learning to optimize it. The supervised method is employed in pre-training to initialize the parameters of $f(\cdot)$. For such pre-training, we assign the ground-truth coordinates $(u_{x}^{i*}, u_{y}^{i*}, u_{l}^{i*})$ for the sample $x_{i}$ as follows: 1) The locations $u_{x}^{i*}$ and $u_{y}^{i*}$ are set to $0.5$ since a chromosome is centered in the image. 2) Based on $u_{x}^{i*}$ and $u_{y}^{i*}$, the smallest region that covers the whole chromosome is calculated and $u_{l}^{i*}$ is computed accordingly. The lower bound of $u_{l}^{i*}$ is $0$ and if the width or height of chromosome exceeds $2T_1$, $u_{l}^{i*}$ will be set to $1$. Given a set of $N$ sample pairs $\{(x_{i}, u_{x}^{i*}, u_{y}^{i*}, u_{l}^{i*})\}_{i=1,2,...,N}$, our loss function for supervised learning is defined as:
\begin{equation}
\begin{split}
\mathcal{L}_{u}(\mathbf{U, U^*}) = \sum^{N}_{i = 1}& \sum_{\gamma \in \{x, y, l\}}\text{smooth}_{L_1}(u_{\gamma}^{i}- u_{\gamma}^{i*}),\\
\text{smooth}_{L_1}(x) &= 
\begin{cases}
0.5x^2        &\text{if}\ |x| < 1\\
|x| - 0.5  &\text{otherwise},
\end{cases}
\end{split}
\label{eq:uloss}
\end{equation}
\noindent where $\mathbf{U}$ and $\mathbf{U^*}$ denote the vector of the predicted coordinates and their ground-truth labels, respectively. The robust $\text{smooth}_{L_1}$ loss \cite{girshick2015fast} is used to directly train the localization subnet to output initial local region coordinates. {It is less sensitive to outliers than $L_2$ loss and smoother near zero compared to the standard $L_1$ norm. Its gradient is uniquely defined at zero point.}

While the weakly-supervised method is aimed at improving the classification performance of the L-Net by optimizing the $f(\cdot)$ for finer part localization, we keep all parameters of the L-Net unchanged and only fine-tune the localization subnet by minimizing the multi-task loss (\ref{eq:weightedloss}) of the L-Net. Without ground-truth coordinates provided, the subnet $f(\cdot)$ autonomously learns to locate discriminative parts, making the extracted features meaningful. Thus, the total loss is given by:
\begin{equation}
\begin{aligned}
\mathcal{L}_{loc}(\mathbf{U, U^*, O^t, Y^t, O^p, Y^p}) = \mathcal{L}_{u}(\mathbf{U, U^*}) + \\
\mathcal{L}_{cls}(\mathbf{O^t, Y^t, O^p, Y^p}).
\end{aligned}
\label{eq:locloss}
\end{equation}
\noindent Here, the subnet is only pre-trained once by minimizing $\mathcal{L}_{u}(\mathbf{U, U^*})$. Then, its optimization process is dominant by weakly-supervised learning. The training details of our proposed Varifocal-Net will be introduced in Sec. \ref{sec:altertrain}.

\subsubsection{Back-propagation through Boxcar Function}
\label{sec:backprop}
We adopt the boxcar function for localization because it provides analytical representations between region cropping and the predicted relative coordinates $(u_x^i, u_y^i, u_l^i)$, which is indispensable for parameter update in back-propagation. When optimizing $\mathcal{L}_{cls}(\mathbf{O^t, Y^t, O^p, Y^p})$ to train the localization subnet, gradients back-propagate through the boxcar function. For one single image $x_i$, we designate the gradients that back-propagate to the input layer of the L-Net as $\mathbf{G}_{top}$. The partial derivatives of the loss to coordinates are then given by:
\begin{equation}
\begin{aligned}
\frac{\partial \mathcal{L}_{cls}(\mathbf{O^t, Y^t, O^p, Y^p})}{\partial u_{\gamma}^{i}} \varpropto \mathbf{G}_{top} \odot \frac{\partial boxcar(t_x^i, t_y^i, t_{l}^i)}{\partial t_{\gamma}^{i}} \cdot \frac{\partial t_{\gamma}^{i}}{\partial u_{\gamma}^{i}},\\
\frac{\partial t_{x}^{i}}{\partial u_{x}^{i}} = \frac{\partial t_{y}^{i}}{\partial u_y^{i}} =  T_2,\ \frac{\partial t_{l}^{i}}{\partial u_{l}^{i}} = T_1/2,\ \gamma \in \{x, y, l\},
\end{aligned}
\label{eq:derivative}
\end{equation}
\noindent where $\odot$ denotes element-wise multiplication. Hence, the derivatives of $boxcar(t_x^i, t_y^i, t_{l}^i)$ with respect to $t_{x}^{i}$, $t_{y}^{i}$ and $t_{l}^{i}$ largely influence the moving direction and size of the local region box. Note that in the context of minimizing our loss, it holds true for $\forall \gamma \in \{x, y, l\}$ that $t_{\gamma}^{i}$ increases when $\frac{\partial \mathcal{L}_{cls}(\mathbf{O^t, Y^t, O^p, Y^p})}{\partial u_{\gamma}^{i}} < 0$ and decreases otherwise. To achieve a consistent optimization direction, we follow \cite{fu2017look} to calculate the negative squared norm of derivatives $\mathbf{G}_{top}$ and compute the $boxcar(t_x^i, t_y^i, t_{l}^i)$'s partial derivatives explicitly in the back-propagation process.

\subsection{Stage 2: Classification based on the Fused Feature}

\begin{figure}[htbp]
\centerline{\includegraphics[scale=0.05]{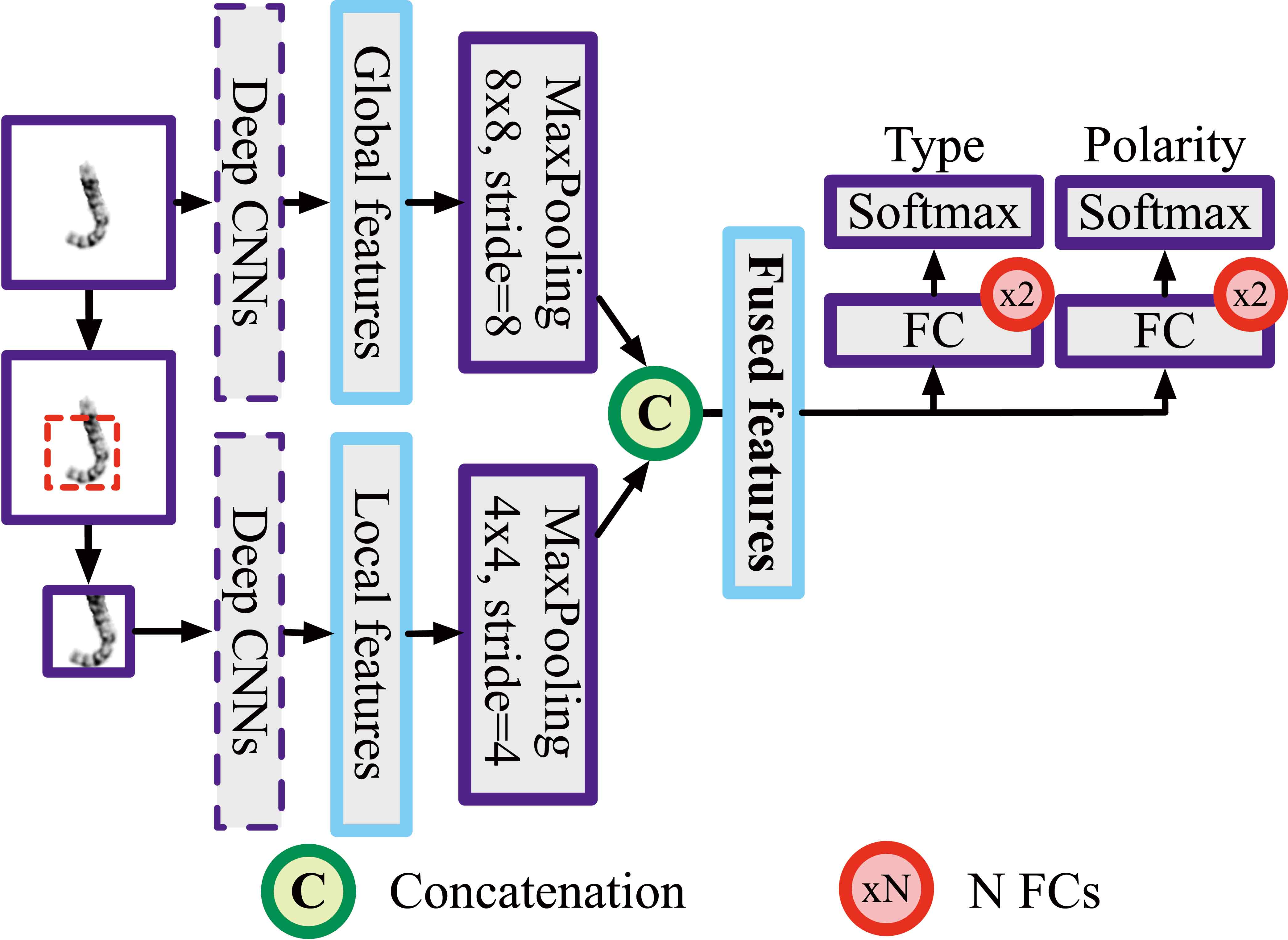}}
\caption{The second stage of the proposed Varifocal-Net: chromosome classification using fused features from both global and local scales.}
\label{totalpredict}
\end{figure}

Once both the G-Net and the L-Net are optimized, global-scale and local-scale features can be extracted via deep CNNs. To make full use of these two representations, it is reasonable to concatenate them into a feature ensemble. We build two MLP classifiers (see Fig. \ref{totalpredict}) to learn the mapping from the fused features to classification probabilities of type and polarity, respectively. Each classifier consists of two FC layers and one Softmax layer. With the trained classifiers, the proposed Varifocal-Net simultaneously predicts chromosome's type and polarity in an end-to-end manner.

\subsection{Four-step Training Strategy}
\label{sec:altertrain}
In this paper, we adopt a four-step optimization technique to alternatively train the network. In the first step, we initialize deep CNNs of the G-Net and L-Net via He's method \cite{he2015delving}. In the second step, we train deep CNNs and the classification subnet in the G-Net until convergence. At this point, the localization subnet and the L-Net are not optimized. In the third step, we prepare all the ground-truth coordinates of local region boxes and {only pre-train the localization subnet once. Finally, we train the L-Net and the localization subnet alternatively in the fourth step}. Keeping the parameters of the localization subnet fixed, we optimize the L-Net by minimizing our multi-task loss. Then we fix the parameters of the L-Net and fine-tune the localization subnet alone. Such alternative training can be run for iterations until {there is no further error loss decrease}.

\subsection{Stage 3: Type Assignment using Dispatch Strategy}
\label{sec:dispatchstrategy}
{In karyotyping practice, the classification of chromosome's type is conducted within each patient case. Therefore the classification can also be viewed as dispatching each chromosome to a certain type. This led us to propose a dispatch strategy for type assignment in the third stage. The design of the dispatch strategy follows two simple rules about karyotyping's domain knowledge \cite{mcgowan2016international}:
\begin{itemize}
\item Each healthy patient has 46 chromosomes for 23 classes (female) or 24 classes (male).
\item For unhealthy patient, the number of each type falls between 1 and 3 (e.g., monosomy 21 and trisomy 21) except extremely rare cases. Type Y has less than 3 chromosomes.
\end{itemize}}

{Considering both the maximum likelihood criterion and possible abnormality situations, we dispatch chromosomes twice. Given the predicted probabilities from the second stage of the Varifocal-Net, the first-time dispatch is to assign each chromosome to the type having the highest probability. The second-time dispatch is to check and compare the probabilities of different chromosomes that are assigned to the same type. The confidence threshold $th$ is designed to filter out uncertain assignments. The dispatch strategy is described in details in Alg. \ref{alg:dispatchstr}. Note that it is not used for polarity prediction because polarity only involves 2 classes (q-arm upward or downward).
\begin{algorithm}[htb]
  \caption{{Dispatch strategy for chromosome's type.}}
  \label{alg:dispatchstr}
  \begin{algorithmic}[1]
    \Require
      $N$ chromosomes; the probabilities of 24 types $P_i$ for the $i$-th chromosome  ($P_{ij}$ stands for its probability of being type $j$, $i=1,2,...,N, j=1,2,...,24$); confidence threshold $th$.
    \Ensure
    The set of chromosomes assigned to type $k$ ($O_k, k=1,2,...,24$); possible abnormal warnings.
    \State $T_k = \varnothing, O_k = \varnothing, \forall k \in\{1,2,...,24\}$.
    \label{code:dispatch:init}
    \For{each $i\in\{1,2,...,N\}$}
    \State Compute the most probable type $j^* = \argmax_{j} P_{ij}$ and dispatch the $i$-th chromosome to type $j^*$ by $T_{j^*} = T_{j^*}\cup \{i\}$;
    \EndFor
    \For {each $k\in\{1,2,...,24\}$}
    {
    \State $S = 1$ if $k =24$, otherwise $S = 2$;
    \If{$\lvert T_k\vert >S$}
    {\State Sort each element in $T_k$ based on its probability. From $T_k$, choose $S+1$ elements ($Q_k=\{i^1,...,i^{S+1}\}$) with the highest probability if $P_{ik} > th, \forall i\in Q_k$, otherwise choose only $S$ elements ($Q_k=\{i^1,...,i^{S}\}$);
    \State $O_k = O_k \cup Q_k$;
    \For {$i\in T_k\setminus Q_k$}
    \State Compute the second probable type $j^* = \argmax_{j, j\neq k} P_{ij}$ and dispatch it to type $j^*$ by $O_{j^*} = O_{j^*} \cup \{i\}$;
    \EndFor
    }
    \Else
    \State $O_k = O_k \cup T_k$;
    \EndIf
    }
    \EndFor
    \State Print abnormal warnings if $\lvert O_k\rvert \neq 2, \forall k \in \{1,2,...,22\}$ or $\lvert O_{23}\rvert +\lvert O_{24}\rvert \neq 2$;\\
    \Return $O_k, k = 1,2,...,24$.
  \end{algorithmic}
\end{algorithm}}

\section{Experiments and Results}
\label{sec:experiment}

\subsection{Materials}
\label{sec:material}
For the experiments conducted in this section, we collected 1909 {different} {patients' karyotyping} cases from the Xiangya Hospital of Central South University, China. {Each patient case contains one Giemsa stained microscopic image of meta-phase chromosomes. All images} are grayscale and sampled with the same resolution, using the Leica's CytoVision System (GSL-120). Each chromosome is of approximate 300-band levels. The datasets contain 1784 karyotyping cases from healthy patients (1061 male and 723 female) and 125 cases from unhealthy patients (73 male and 52 female). The unhealthy cases contain both numerical and structural abnormalities. Each chromosome's type is manually annotated by cytogeneticists in real-world clinical environments. The type of autosomes is labeled from 0 to 21 and the type of sex chromosomes X and Y are denoted as 22 and 23, respectively. The polarity of a chromosome is labeled as 1 if its q-arm is downward and 0 otherwise.

\begin{table}[ht]
\centering
\caption{{Statistics of the dataset. (H: Healthy Samples, U: Unhealthy Samples.)}}
\label{tab:dataset}
\begin{tabular}{lccccccc}
\hline
\multicolumn{2}{c}{\multirow{2}{*}{{}Dataset}} & \multicolumn{2}{c}{Case \#} &  & \multicolumn{2}{c}{Image \#} & \multirow{2}{*}{\begin{tabular}[c]{@{}c@{}}Total\\ image \#\end{tabular}} \\ \cline{3-4} \cline{6-7}
\multicolumn{2}{c}{} & Male & Female &  & Male & Female &  \\ \hline
\multirow{2}{*}{\begin{tabular}[c]{@{}c@{}}Total\\ samples\end{tabular}} & H & 1061 & 723 &  & 48806 & 33258 & \multirow{2}{*}{87831} \\
 & U & 73 & 52 &  & 3384 & 2383 &  \\ \hline
\end{tabular}
\end{table}

We obtain each individual chromosome image by manually segmenting it from {microscopic images}. In total, there exist 87831 separated chromosomes. {We randomly split both healthy and unhealthy samples into five subsets to perform five-fold cross validation. Each time, four subsets are used for training the model and fine-tuning the hyper-parameters. The remaining one subset is left for testing. Note that the chromosome samples are divided by patient case. All chromosomes of the same case stay in the same subset.} Table \ref{tab:dataset} provides the details of our datasets.

\subsection{Implementation Details}
\label{sec:imdetail}

The size of images differs from each other and we first padded them with pixels into square images of the same size. The padding value is set as $255$ to imitate the background of the original Giemsa stained images. And the size of padded image is $320\times320$ pixels. Then, we resized the image to $256\times256$ pixels and normalized all $N$ images as follows:
\begin{equation}
\begin{aligned}
x_i' = (x_i - \mu_i)/\sigma_i,\ i = 1,2,...,N
\end{aligned}
\label{eq:preprocess}
\end{equation}
\noindent where $\mu_i$ and $\sigma_i$ are the mean value and the standard deviation of the sample $x_i$, respectively. $x_i'$ denotes the normalized input, which has a zero mean and a unit variance. For local region prediction, the margin $T_1$ is $64$ and the shift range $T_2$ is $128$. The cropped target region was then upsampled to $128\times128$ pixels as  the input to the L-Net. In Table \ref{table:fc1} and Table \ref{table:fc2}, we describe feature dimensions of the proposed Varifocal-Net for the first and the second stages, respectively. {For the dispatch strategy, the confidence threshold $th$ was set to $0.9$ because we only keep highly-confident chromosomes when possible numerical abnormalities happen.}

\begin{table}[htbp]
\centering
\caption{The feature dimensions of the Varifocal-Net for the first stage. (T: Type, P: Polarity, Loc: Localization.)}
\label{table:fc1}
\begin{tabular}{cccccc}
\hline
\multirow{2}{*}{Layer} & \multicolumn{5}{c}{Dimension} \\ \cline{2-6} 
 & \multicolumn{3}{c}{G-Net} & \multicolumn{2}{c}{L-Net} \\ \hline
Input & \multicolumn{3}{c}{$256\times256$} & \multicolumn{2}{c}{$128\times128$} \\
Deep CNNs & \multicolumn{3}{c}{$640\times32\times32$} & \multicolumn{2}{c}{$640\times16\times16$} \\
Max-pooling & \multicolumn{2}{c}{$640\times4\times4$} & $640\times8\times8$ & \multicolumn{2}{c}{$640\times4\times4$} \\
FC1 & $24$ (T) & $2$ (P) & $1024$ & $24$ (T) & $2$ (P) \\
FC2 & $-$ & $-$ & $3$ (Loc) & $-$ & $-$ \\ \hline
\end{tabular}
\end{table}

\begin{table}[htbp]
\centering
\caption{The feature dimensions of the Varifocal-Net for the second stage. (T: Type, P: Polarity.)}
\label{table:fc2}
\begin{tabular}{cclcl}
\hline
Layer & \multicolumn{4}{c}{Dimension} \\ \hline
Input & \multicolumn{2}{c}{\begin{tabular}[c]{@{}c@{}}$256\times256$ (G-Net)\end{tabular}} & \multicolumn{2}{c}{\begin{tabular}[c]{@{}c@{}}$128\times128$ (L-Net)\end{tabular}} \\
Deep CNNs & \multicolumn{2}{c}{\begin{tabular}[c]{@{}c@{}}$640\times32\times32$ \end{tabular}} & \multicolumn{2}{c}{\begin{tabular}[c]{@{}c@{}}$640\times16\times16$ \end{tabular}} \\
Max-pooling & \multicolumn{2}{c}{$640\times4\times4$} & \multicolumn{2}{c}{$640\times4\times4$} \\
Concatenation & \multicolumn{4}{c}{$640\times4\times4\times2$} \\
FC1 & \multicolumn{2}{c}{$512$} & \multicolumn{2}{c}{$512$} \\
FC2 & \multicolumn{2}{c}{$24$ (T)} & \multicolumn{2}{c}{$2$ (P)} \\ \hline
\end{tabular}
\end{table}

In the training process, we adopted horizontal flipping and random rotation between $[0^{\circ}, 45^{\circ}]$ for data augmentation. The vertical flipping operation was performed to change the polarity label of a chromosome. All modules of the Varifocal-Net were trained from scratch using Adam optimizer \cite{kingma2014adam} with $\beta_1 = 0.9$ and $\beta_2=0.999$. The initial learning rate was set to $0.0001$ and it decreased by nine-tenth every 10 epochs. We implemented the proposed Varifocal-Net and other CNN-based methods in Python, with PyTorch framework \cite{paszke2017automatic}. All experiments were conducted under a Ubuntu OS workstation with Intel Xeon(R) CPU E5-2620 v4 $@$ 2.10GHz, 128 GB of RAM, and 4 NVIDIA GTX Titan X GPUs.

\subsection{Evaluation Metrics}
\label{sec:metric}
{The performance of the Varifocal-Net was evaluated by four metrics: the accuracy of all the testing images (Acc.), the average $F_1$-score over classes of all the testing images ($F_1$), the average accuracy of the complete karyotyping per patient case (Acc. per Case), and the average accuracy of the complete karyotyping per patient case using the proposed dispatch strategy (Acc. per Case-D). The Acc.} is an intuitive measurement defined as the fraction of the testing samples which are correctly classified.

For the computation of $F_1$-score, we first define the following four criteria to fit the context of multi-class classification:
\begin{itemize}
\item True positives ($TP_j$): images predicted as class $j$ which actually belong to class $j$
\item False positives ($FP_j$): images predicted as class $j$ which actually do not belong to class $j$
\item False negatives ($FN_j$): images predicted as class $k$ $(\forall k\neq j)$ which actually belong to class $j$
\item True negatives ($TN_j$): images predicted as class $k$ $(\forall k\neq j)$ which actually do not belong to class $j$
\end{itemize}

Then, the $F_1$-score is computed as:
\begin{equation}
\begin{split}
F_1 = \frac{1}{N_{cls}}\sum^{N_{cls}}_{j=1}& \frac{2\cdot Precision_{j}\cdot Recall_{j}}{Precision_{j}+Recall_{j}}, \\
Precision_{j} &= \frac{TP_j}{TP_j+FP_j},\\
Recall_{j} &= \frac{TP_j}{TP_j+FN_j},
\end{split}
\label{eq:f1score}
\end{equation}
\noindent where $N_{cls}$ equals 24 and 2 for type and polarity recognition, respectively.

{The accuracy per patient case was adopted to evaluate the performance in clinical settings. It is computed by checking the fraction of the correctly classified samples within each patient case. No dispatch strategy is used for computing Acc. per Case. We only assign each chromosome to the type having the highest predicted probability. For the computation of Acc. per Case-D, the proposed dispatch strategy is employed and accuracy within each case is recalculated for all samples.}

{The mean value and the standard deviation of these four metrics are provided to assess performance stability. They were calculated based on the results of five-fold cross validation and displayed in percentage.}

Furthermore, we also adopted a receiver operating characteristic (ROC) analysis for performance comparison. The ROC curves averaged over all classes were plotted and the area under each curve (AUC) was calculated as well.

\subsection{Results}
\label{sec:results}
This section presents experimental results in three parts. We first provide detailed evaluation results of the proposed Varifocal-Net. Then, a comparison of the proposed method with state-of-the-art methods is given. Finally, we present additional results for analyzing our performance.

\subsubsection{Evaluation Results}
\label{sec:evalresults}
\begin{table*}[!t]
\centering
\caption{{Performance of the Varifocal-Net (mean$\pm$standard deviation). The results are presented in terms of four evaluation metrics: average $F_1$-score of all testing images ($F_1$), accuracy of all testing images (Acc.), average accuracy per patient case (Acc. per Case), and average accuracy per patient case using the proposed dispatch strategy (Acc. per Case-D). (T: Type, P: Polarity, PET: Per Epoch Time, TPI: Time Per Image.)}}
\label{table:evaluation}
\begin{tabular}{ccccccccccc}
\hline
\multirow{2}{*}{{}Stage} & \multirow{2}{*}{Method} & \multicolumn{2}{c}{$F_1$ (\%)} & \multicolumn{2}{c}{Acc. (\%)} & \multicolumn{2}{c}{\begin{tabular}[c]{@{}c@{}} Acc. per\\Case (\%)\end{tabular}} & \begin{tabular}[c]{@{}c@{}} Acc. per\\Case-D (\%)\end{tabular} & \multirow{2}{*}{\begin{tabular}[c]{@{}c@{}}\# Epoch \\$\times$PET (s)\end{tabular}} & \multirow{2}{*}{\begin{tabular}[c]{@{}c@{}}Testing\\TPI (ms)\end{tabular}} \\  \cline{3-9}
 &  & T & P & T & P & T & P & T &  &  \\ \hline
\multirow{2}{*}{1} & G-Net & 97.5$\pm$0.4 & 99.0$\pm$0.1 & 97.8$\pm$0.4 & 99.0$\pm$0.1 &97.8$\pm$3.8  &99.0$\pm$1.9  &98.2$\pm$3.3  &30$\times$956.3$\pm$1.5  &5.7$\pm$0.1  \\
 & L-Net & 98.2$\pm$0.5 & 99.2$\pm$0.1 & 98.4$\pm$0.5 & 99.2$\pm$0.1  &98.4$\pm$2.9  &99.2$\pm$1.6  &98.9$\pm$2.5 &30$\times$1142.3$\pm$2.3  &6.8$\pm$0.1  \\ \hline
2 & \textbf{Varifocal-Net} & \textbf{98.7$\pm$0.7} & \textbf{99.2$\pm$0.3} & \textbf{98.9$\pm$0.7} & \textbf{99.2$\pm$0.3} &\textbf{98.9$\pm$2.3} &\textbf{99.2$\pm$1.5}  &\textbf{99.2$\pm$2.1}  &\textbf{20$\times$1150.8$\pm$11.3} &\textbf{5.9$\pm$0.1}  \\ \hline
\end{tabular}
\end{table*}

Table \ref{table:evaluation} gives the classification results of the G-Net, L-Net, and the entire Varifocal-Net. The global-scale G-Net achieved the {accuracy (\%) of 97.8 and 99.0} for type and polarity recognition, respectively. With the localization subnet for finer region detection, the local-scale L-Net reduced classification errors. By utilizing the knowledge learned at two scales, the proposed Varifocal-Net yielded {the best} performance. The accuracy (\%) of type and polarity tasks were boosted to {98.9 and 99.2}, respectively. {Due to the proposed dispatch strategy, the accuracy of type classification per case is further improved for each method. The proposed Varifocal-Net achieved the averaged Acc. per Case-D (\%) of 99.2.} Though the total training time is relatively long, the testing time of the Varifocal-Net is only {5.9ms} per sample.

\begin{table}[htbp]
\centering
\caption{{Performance of the Varifocal-Net for each chromosome type (mean$\pm$standard deviation).}}
\label{table:alltype}
\begin{tabular}{cccc}
\hline
Class (No.) & $F_1$ (\%) & Precision (\%) & Recall (\%) \\ \hline
1 &99.6$\pm$0.6  &99.5$\pm$0.7  &99.7$\pm$0.5  \\
2 &99.3$\pm$0.7  &98.8$\pm$0.9  &99.7$\pm$0.5  \\
3 &99.5$\pm$0.6  &99.4$\pm$0.8  &99.6$\pm$0.5  \\
4 &98.6$\pm$1.1  &98.4$\pm$1.1  &98.7$\pm$1.1  \\
5 &98.6$\pm$0.7  &98.7$\pm$0.7  &98.6$\pm$0.7  \\
6 &99.4$\pm$0.6  &99.7$\pm$0.3  &99.2$\pm$0.8  \\
7 &99.7$\pm$0.2  &99.7$\pm$0.3  &99.6$\pm$0.3  \\
8 &98.9$\pm$0.8  &98.9$\pm$0.9  &98.9$\pm$0.8  \\
9 &98.7$\pm$0.4  &98.8$\pm$0.8  &98.6$\pm$0.5  \\
10 &98.7$\pm$0.7  &98.7$\pm$0.8  &98.7$\pm$0.7  \\
11 &99.6$\pm$0.2  &99.6$\pm$0.3  &99.6$\pm$0.3  \\
12 &99.7$\pm$0.2  &99.8$\pm$0.1  &99.6$\pm$0.4  \\
13 &98.7$\pm$0.7  &98.8$\pm$0.5  &98.7$\pm$1.0  \\
14 &99.0$\pm$0.5  &99.2$\pm$0.6  &98.9$\pm$0.5  \\
15 &98.5$\pm$0.8  &98.6$\pm$0.9  &98.4$\pm$0.7  \\
16 &97.9$\pm$1.3  &97.9$\pm$1.2  &97.9$\pm$1.4  \\
17 &99.3$\pm$0.6  &99.1$\pm$0.8  &99.4$\pm$0.4  \\
18 &98.8$\pm$1.1  &98.9$\pm$0.8  &98.7$\pm$1.5  \\
19 &98.7$\pm$0.8  &98.6$\pm$0.9  &98.8$\pm$0.8  \\
20 &98.4$\pm$1.0  &98.5$\pm$1.1  &98.4$\pm$0.9  \\
21 &98.5$\pm$0.5  &98.5$\pm$0.4  &98.6$\pm$0.7  \\
22 &98.4$\pm$0.8  &98.3$\pm$0.8  &98.6$\pm$0.9  \\
X &98.3$\pm$1.1  &98.6$\pm$0.9  &98.1$\pm$1.4  \\
Y &94.3$\pm$3.6  &95.0$\pm$3.5  &93.6$\pm$3.8  \\ \hline
\end{tabular}
\end{table}

\begin{table}[htbp]
\centering
\caption{{Performance of the Varifocal-Net for each chromosome polarity (mean$\pm$standard deviation).}}
\label{table:allpolarity}
\begin{tabular}{cccc}
\hline
Class & $F_1$ (\%) & Precision (\%)& Recall (\%)\\ \hline
q-arm upward &99.2$\pm$0.3   &99.1$\pm$0.4   &99.4$\pm$0.1  \\
q-arm downward &99.3$\pm$0.3   &99.4$\pm$0.1   &99.1$\pm$0.4  \\ \hline
\end{tabular}
\end{table}

To observe the performance of the Varifocal-Net on each class of chromosomes, Table \ref{table:alltype} and Table \ref{table:allpolarity} provide the $F_1$-score, precision, and recall, which were computed within each category. For type recognition, the proposed method performed worst on Y chromosomes, with only a $F_1$-score (\%) of {94.3} achieved. The evaluation results of classes {No. 4, No. 5}, No.15, No. 16, No. 20--No. 22, X, and Y are below average. For polarity recognition, the orientation of q-arm was accurately predicted, with the $F_1$-score of each class above {99\%}.

{Besides, for polarity classification,} we also computed the accuracy within each type category to learn the performance difference among chromosome types. Table \ref{table:upclass} {indicates} that our prediction is relatively inaccurate for two long types ({classes No. 2 and No. 5}) and {four} short types (classes No. 15, {No.16}, No. 20 and Y).

\begin{table}[htbp]
\centering
\caption{{Performance of the Varifocal-Net for polarity classification within each type (mean$\pm$standard deviation).}}
\label{table:upclass}
\begin{tabular}{cccccc}
\hline
\begin{tabular}[c]{@{}c@{}}Class\\(No.)\end{tabular}  & Acc. (\%) & \begin{tabular}[c]{@{}c@{}}Class\\(No.)\end{tabular} & Acc. (\%) & \begin{tabular}[c]{@{}c@{}}Class\\(No.)\end{tabular} & Acc. (\%) \\ \hline
1 &99.3$\pm$0.5  & 9 &99.6$\pm$0.1  & 17 &99.3$\pm$0.5 \\
2 &99.1$\pm$0.5  & 10 &99.5$\pm$0.2 & 18 &99.5$\pm$0.1 \\
3 &99.5$\pm$0.4 & 11 &99.8$\pm$0.2  & 19 &99.3$\pm$0.4 \\
4 &99.2$\pm$0.4  & 12 &99.6$\pm$0.2 & 20 &96.2$\pm$1.1    \\
5 &99.1$\pm$0.3 & 13 &99.6$\pm$0.4 & 21 &99.3$\pm$0.3   \\
6 &99.5$\pm$0.2 & 14 &99.8$\pm$0.2  & 22 &99.5$\pm$0.1   \\
7 &99.8$\pm$0.2 & 15 &98.9$\pm$0.5  & X &99.2$\pm$0.3  \\
8 &99.5$\pm$0.2  & 16 &99.1$\pm$0.4  & Y &98.0$\pm$0.8 \\ \hline
\end{tabular}
\end{table}

\subsubsection{Comparison with the State-of-the-Art}

Table \ref{table:comp2} provides a comparison of the proposed Varifocal-Net with state-of-the-art methods. {The first two methods \cite{sharma2017crowdsourcing, gupta2017siamese} were proposed specifically for classifying Giemsa stained chromosomes. Both the two existing methods employed CNNs for feature extraction, and} they relied on straightening chromosomes for normalization and used small datasets. In contrast, we adopted an end-to-end fashion for prediction. We implemented the two methods and {evaluated them using five-fold cross validation}. Their performance {of type recognition} on our large testing set proves the superiority of our method, which surpasses \cite{sharma2017crowdsourcing} and \cite{gupta2017siamese} by {nearly 6.7\% and 7.5\% in average $F_1$-score, respectively.}

\begin{table*}[htbp]
\centering
\caption{{Comparison results of the proposed method with state-of-the-art methods (mean$\pm$standard deviation). The results are presented in terms of four evaluation metrics: average $F_1$-score of all testing images ($F_1$), accuracy of all testing images (Acc.), average accuracy per patient case (Acc. per Case), and average accuracy per patient case using the proposed dispatch strategy (Acc. per Case-D). (T: Type, P: Polarity.)}}
\label{table:comp2}
\begin{tabular}{lcccccccc}
\hline
\multirow{2}{*}{{}Method} & \multicolumn{2}{c}{$F_1$ (\%)} & \multicolumn{2}{c}{Acc. (\%)} & \multicolumn{2}{c}{Acc. per Case (\%)} & Acc. per Case-D (\%) \\ \cline{2-8} 
 & T & P & T & P & T & P & T \\ \hline
Sharma \textit{et al.} \cite{sharma2017crowdsourcing}& 92.0$\pm$1.6 & -- & 92.6$\pm$1.5 & -- & 92.6$\pm$7.9  & -- & 93.6$\pm$7.4 \\
Gupta \textit{et al.} \cite{gupta2017siamese}& 91.2$\pm$2.3 & -- & 91.8$\pm$2.2 & -- & 91.8$\pm$9.9  & -- & 92.6$\pm$9.5 \\ \hline
AlexNet \cite{krizhevsky2012imagenet} & 90.2$\pm$1.9 & 97.1$\pm$0.5 & 90.8$\pm$1.8 & 97.1$\pm$0.5 & 90.8$\pm$9.5  & 97.1$\pm$3.9 & 92.4$\pm$9.1 \\
GoogLeNet \cite{szegedy2016rethinking} &95.6$\pm$1.6  &98.6$\pm$0.5  &96.0$\pm$1.5  &98.6$\pm$0.5 &96.0$\pm$6.5  &98.6$\pm$2.7  &96.8$\pm$6.2  \\
VGG-Net \cite{simonyan2014very} & 96.0$\pm$0.7 & 98.8$\pm$0.2 & 96.3$\pm$0.6 & 98.8$\pm$0.2  &96.3$\pm$5.3  &98.8$\pm$2.2  &97.1$\pm$4.9 \\
ResNet \cite{he2016deep} & 96.6$\pm$0.9 & 98.9$\pm$0.2 & 96.9$\pm$0.9 & 98.9$\pm$0.2 &96.9$\pm$4.7  &98.9$\pm$2.1  &97.5$\pm$4.2  \\
DenseNet \cite{huang2017densely} & 96.2$\pm$1.3 & 98.8$\pm$0.4 & 96.5$\pm$1.2 & 98.8$\pm$0.4 &96.5$\pm$5.4  &98.8$\pm$2.2  &97.3$\pm$4.9  \\
AlexNet-STN \cite{krizhevsky2012imagenet,jaderberg2015spatial} &92.9$\pm$2.1  &97.8$\pm$0.5  &93.4$\pm$2.0  &97.8$\pm$0.5  &93.4$\pm$7.4  &97.8$\pm$3.3  &94.7$\pm$6.9 \\
GoogLeNet-STN \cite{szegedy2016rethinking,jaderberg2015spatial} &90.7$\pm$1.8  &97.4$\pm$0.3  &91.2$\pm$1.8  &97.4$\pm$0.3 &91.2$\pm$9.8  &97.4$\pm$3.8  &93.1$\pm$9.5  \\
VGG-Net-STN \cite{simonyan2014very,jaderberg2015spatial} &96.8$\pm$0.8  &99.0$\pm$0.3  &97.1$\pm$0.8  &99.0$\pm$0.3  &97.0$\pm$4.4  &99.0$\pm$1.9  &97.7$\pm$4.1  \\
ResNet-STN \cite{he2016deep,jaderberg2015spatial} &96.9$\pm$0.9  &98.9$\pm$0.2  &97.2$\pm$0.9  &98.9$\pm$0.2  &97.2$\pm$3.7  &98.9$\pm$1.8  &97.8$\pm$3.3  \\
DenseNet-STN \cite{huang2017densely,jaderberg2015spatial} &97.0$\pm$1.5  &99.0$\pm$0.4  &97.3$\pm$1.4  &99.0$\pm$0.3  &97.3$\pm$4.4  &99.0$\pm$1.8  &97.9$\pm$3.9  \\ \hline
G-Net-STN \cite{jaderberg2015spatial}&95.9$\pm$1.8  &98.7$\pm$0.4  &96.2$\pm$1.6  &98.7$\pm$0.4  &96.2$\pm$5.6  &98.7$\pm$2.2  &97.2$\pm$5.0  \\
L-Net (Simple)& 95.3$\pm$0.8  & 98.3$\pm$0.4   & 95.8$\pm$0.7  & 98.3$\pm$0.4  & 95.8$\pm$4.9   & 98.3$\pm$2.5 & 97.0$\pm$4.1   \\
Varifocal-Net (Simple)& 96.1$\pm$0.6  & 98.3$\pm$0.2   & 96.5$\pm$0.5   & 98.3$\pm$0.2   & 96.5$\pm$4.7   & 98.3$\pm$2.4    & 96.6$\pm$4.7    \\ \hline
\textbf{Varifocal-Net} & \textbf{98.7$\pm$0.7} & \textbf{99.2$\pm$0.3} & \textbf{98.9$\pm$0.7} & \textbf{99.2$\pm$0.3} &\textbf{98.9$\pm$2.3} &\textbf{99.2$\pm$1.5}  &\textbf{99.2$\pm$2.1}  \\ \hline
\end{tabular}
\end{table*}

{To test the usefulness of the varifocal mechanism, we replaced the localization subnet by a simple preprocessing method. The input of the L-Net is not the cropped local region of the original image. Instead, we directly rescaled and padded the minimum bounding box of each chromosome image into the same size (256$\times$256). The processed image contains the whole chromosome part and consequently the extracted features are no longer local. After the L-Net converges, the features learned from the G-Net and the L-Net are concatenated as well for the training of the second stage. We named the L-Net and the Varifocal-Net using such a simple preprocessing step as L-Net (Simple) and Varifocal-Net (Simple), respectively. Table \ref{table:comp2} shows that the simple preprocessing method does not facilitate feature learning of fine-grained details. Our method outperforms the L-Net (Simple) and the Varifocal-Net (Simple), which validates the effectiveness of the localization subnet.}

{Table \ref{table:comp2} also provides the results of} comparison with other CNN models. To {assess our multi-scale feature ensemble strategy}, we evaluated the performance of the well-known models that have been proved powerful on the ImageNet dataset, including AlexNet \cite{krizhevsky2012imagenet}, GoogLeNet \cite{szegedy2016rethinking}, VGG-Net-D \cite{simonyan2014very}, ResNet-101 \cite{he2016deep}, and DenseNet-121 \cite{huang2017densely}. The number of convolution layers in these five models and our Varifocal-Net (feature extractor part) are respectively 5, 22, 13, 100, 120, and 28, which are much deeper than previous work in chromosome classification \cite{sharma2017crowdsourcing, gupta2017siamese}. {Besides, we also evaluated Spatial Transformer Network (STN) \cite{jaderberg2015spatial} for performance comparison. It contains 4 Conv layers, 4 Max-Pooling layers, and 2 FC layers. We inserted STN into the first layer of each model and retrained it. Since the parameters of these popular models were compatible with the 3-channel $224\times 224$ natural images (ImageNet)}, we rescaled our $256\times 256$ grayscale images into $224\times 224$ pixels and then generated 3 channels by directly stacking the original grayscale channel. The preprocessing step was also adopted to normalize all the inputs as mentioned in Sec. \ref{sec:imdetail}. To introduce multi-task learning, we duplicated the classifier settings in each model so that both type and polarity could be predicted at the same time. The loss function is defined as (\ref{eq:weightedloss}) with $\lambda = 0.5$. We trained all models from scratch because the collected samples are sufficient. The results show that all models have acceptable performance. Even the shallowest AlexNet achieved the accuracy {(\%) of 90.8 and 97.1} for type and polarity classifications, respectively. Among these single-scale CNN models, the highest accuracy and $F_1$-score were achieved by {DenseNet-STN for both type and polarity tasks}. However, its result is still inferior to ours, {where} the error rates {of type classification} are reduced by half. For the polarity {task}, our method also outperformed other CNN models. {Note that the use of STN does not necessarily improve the performance. Its introduction in GoogLeNet and G-Net brings about obvious decrease.}

\begin{table*}[htbp]
\centering
\caption{{Comparison results of the proposed method with state-of-the-art methods on unhealthy cases (mean$\pm$standard deviation). The results are presented in terms of four evaluation metrics: average $F_1$-score of all testing images ($F_1$), accuracy of all testing images (Acc.), average accuracy per patient case (Acc. per Case), and average accuracy per patient case using the proposed dispatch strategy (Acc. per Case-D). (T: Type, P: Polarity.)}}
\label{table:comp3}
\begin{tabular}{lccccccc}
\hline
\multirow{2}{*}{{}Method} & \multicolumn{2}{c}{$F_1$ (\%)} & \multicolumn{2}{c}{Acc. (\%)} & \multicolumn{2}{c}{Acc. per Case (\%)} & Acc. per Case-D (\%) \\ \cline{2-8} 
 & T & P & T & P & T & P & T \\ \hline
Sharma \textit{et al.} \cite{sharma2017crowdsourcing}& 88.4$\pm$3.5 & -- & 88.9$\pm$3.5 & -- & 88.9$\pm$12.9  & -- & 90.3$\pm$12.5 \\
Gupta \textit{et al.} \cite{gupta2017siamese}& 90.6$\pm$1.0 & -- & 90.8$\pm$1.0 & -- & 90.8$\pm$14.7  & -- & 92.3$\pm$14.1 \\ \hline
AlexNet \cite{krizhevsky2012imagenet} & 80.7$\pm$5.2 & 93.8$\pm$2.2 & 81.1$\pm$5.3 & 94.0$\pm$2.1 & 81.2$\pm$18.6  & 94.0$\pm$7.1 & 83.6$\pm$18.2  \\
GoogLeNet \cite{szegedy2016rethinking} &89.0$\pm$4.8  &96.1$\pm$2.2  &89.3$\pm$4.7  &96.1$\pm$2.2 &89.2$\pm$16.6  &96.1$\pm$7.5  &90.5$\pm$16.2   \\
VGG-Net \cite{simonyan2014very} & 92.1$\pm$2.4 & 97.6$\pm$1.1 & 92.5$\pm$2.5 & 97.6$\pm$1.1 &92.5$\pm$10.8  &97.6$\pm$4.0  &93.7$\pm$10.0  \\
ResNet \cite{he2016deep} & 93.5$\pm$1.7 & 98.0$\pm$1.2 & 93.8$\pm$1.8 & 98.0$\pm$1.2 &93.8$\pm$9.7  &98.0$\pm$3.8  &94.7$\pm$9.0   \\
DenseNet \cite{huang2017densely} & 92.3$\pm$2.8 & 97.7$\pm$1.0 & 92.7$\pm$2.7 & 97.8$\pm$1.0 &92.6$\pm$11.4  &97.8$\pm$4.0  &93.7$\pm$10.9  \\
AlexNet-STN \cite{krizhevsky2012imagenet,jaderberg2015spatial} &87.2$\pm$5.5  &95.7$\pm$2.1  &87.6$\pm$5.6  &95.8$\pm$2.1  &87.5$\pm$15.1  &95.8$\pm$6.1  &89.2$\pm$15.1  \\
GoogLeNet-STN \cite{szegedy2016rethinking,jaderberg2015spatial} &81.0$\pm$4.4  &94.2$\pm$1.8 &81.5$\pm$4.6  &94.3$\pm$1.8 &81.5$\pm$18.7  &94.2$\pm$8.6  &83.4$\pm$19.6  \\
VGG-Net-STN \cite{simonyan2014very,jaderberg2015spatial} &94.4$\pm$2.6  &98.4$\pm$0.8  &94.7$\pm$2.5  &98.4$\pm$0.8  &94.7$\pm$8.3  &98.5$\pm$2.9  &95.7$\pm$8.2  \\
ResNet-STN \cite{he2016deep,jaderberg2015spatial} &96.0$\pm$0.5  &98.6$\pm$0.7  &96.2$\pm$0.5  &98.6$\pm$0.6  &96.2$\pm$5.6  &98.6$\pm$2.7  &96.8$\pm$5.0  \\
DenseNet-STN \cite{huang2017densely,jaderberg2015spatial} &95.8$\pm$2.9  &98.5$\pm$0.9  &96.0$\pm$2.8  &98.5$\pm$0.9  &96.0$\pm$6.7  &98.5$\pm$2.4  &96.7$\pm$6.0 \\ \hline
G-Net   & 95.4$\pm$1.4   & 98.1$\pm$0.8   & 95.6$\pm$1.4   & 98.1$\pm$0.7   & 95.6$\pm$7.3   & 98.1$\pm$3.5   & 96.3$\pm$6.6    \\
L-Net   & 96.6$\pm$1.3   & 98.5$\pm$0.5   &96.8$\pm$1.3   & 98.6$\pm$0.5   & 96.8$\pm$5.8   & 98.6$\pm$2.4   & 97.5$\pm$5.2  \\
G-Net-STN \cite{jaderberg2015spatial}&93.4$\pm$3.2  &97.8$\pm$1.2  &93.6$\pm$3.0  &97.8$\pm$1.2  &93.6$\pm$9.3  &97.8$\pm$3.5  &94.7$\pm$8.8   \\
L-Net (Simple)& 94.5$\pm$1.7  & 97.7$\pm$0.8   & 94.8$\pm$1.6   & 97.8$\pm$0.8  & 94.8$\pm$6.7   & 97.8$\pm$3.0   & 95.8$\pm$6.6   \\
Varifocal-Net (Simple)& 95.1$\pm$1.1  & 97.5$\pm$0.4   & 95.3$\pm$0.9   & 97.5$\pm$0.4  & 95.3$\pm$5.3   & 97.5$\pm$3.1   & 95.4$\pm$5.3   \\ \hline
\textbf{Varifocal-Net}  & \textbf{97.7$\pm$1.6} & \textbf{98.6$\pm$0.6} & \textbf{97.8$\pm$1.7} & \textbf{98.6$\pm$0.6}&\textbf{97.8$\pm$4.3}  &\textbf{98.6$\pm$2.5}  &\textbf{98.4$\pm$3.9} \\ \hline
\end{tabular}
\end{table*}

In the real clinical environment, it is imperative to correctly classify chromosomes having numerical and structural anomalies. To test the robustness of different methods under abnormal circumstance, we specially provide the evaluation results only on unhealthy cases in Table \ref{table:comp3}. For most CNN-based methods, the performance degraded dramatically on abnormal cases. {The AlexNet and GoogLeNet-STN even suffered over 9\% loss of accuracy and $F_1$-score. In contrast,} our Varifocal-Net had only a slight performance drop around {1.1\% and 0.6\% in Acc. per Case of the type and polarity task, respectively.} We remarkably outperformed state-of-the-art methods on abnormal chromosome classification.

\begin{figure}[htbp]
\centerline{\includegraphics[width=\columnwidth]{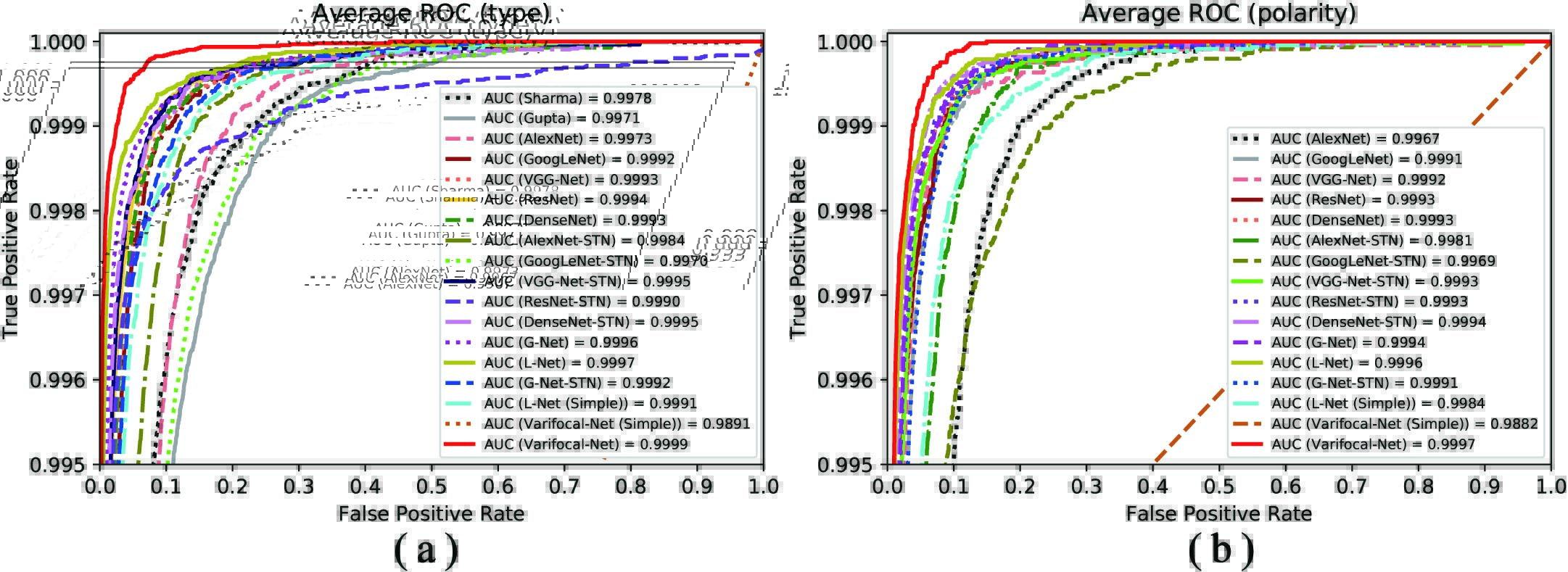}}
\caption{ROC analysis for the proposed Varifocal-Net and previous CNN models. Each ROC is averaged over all classes and its AUC is calculated. (a) ROC of type classification. (b) ROC of polarity classification.}
\label{fig:roc}
\end{figure}

{In Fig. \ref{fig:roc}, the results of ROC analysis are illustrated for both type and polarity classifications. We first performed ROC analysis per class using a one-vs-all scheme. Then, we averaged all ROC curves over classes and calculated the AUC for each method. It is observed} that the proposed Varifocal-Net outperformed other methods with the least false positive predictions and the highest true positive rates. We achieved the highest AUC {for both the type and polarity tasks. It} demonstrates that in the case of not redesigning a completely brand-new feature extraction architecture, our Varifocal-Net, which benefited from the global and local feature ensemble, could further boost the overall classification performance. {The lowest three AUCs of the type task were observed for \cite{krizhevsky2012imagenet}, \cite{sharma2017crowdsourcing}, \cite{gupta2017siamese}, and simple processing methods,} which is consistent with Table \ref{table:comp2}. {Furthermore, statistical tests were performed using both unpaired and paired t-tests \cite{samuels2003statistics, hsu2014paired}. The Acc. per Case of all five fold testing samples were tested and the results of two t-tests confirm the significant superiority of the proposed Varifocal-Net against all other methods (p-value $\ll$ 0.05) for both type and polarity tasks.}

\subsubsection{Performance Analysis Results}
\begin{figure}[htbp]
\centerline{\includegraphics[scale=0.45]{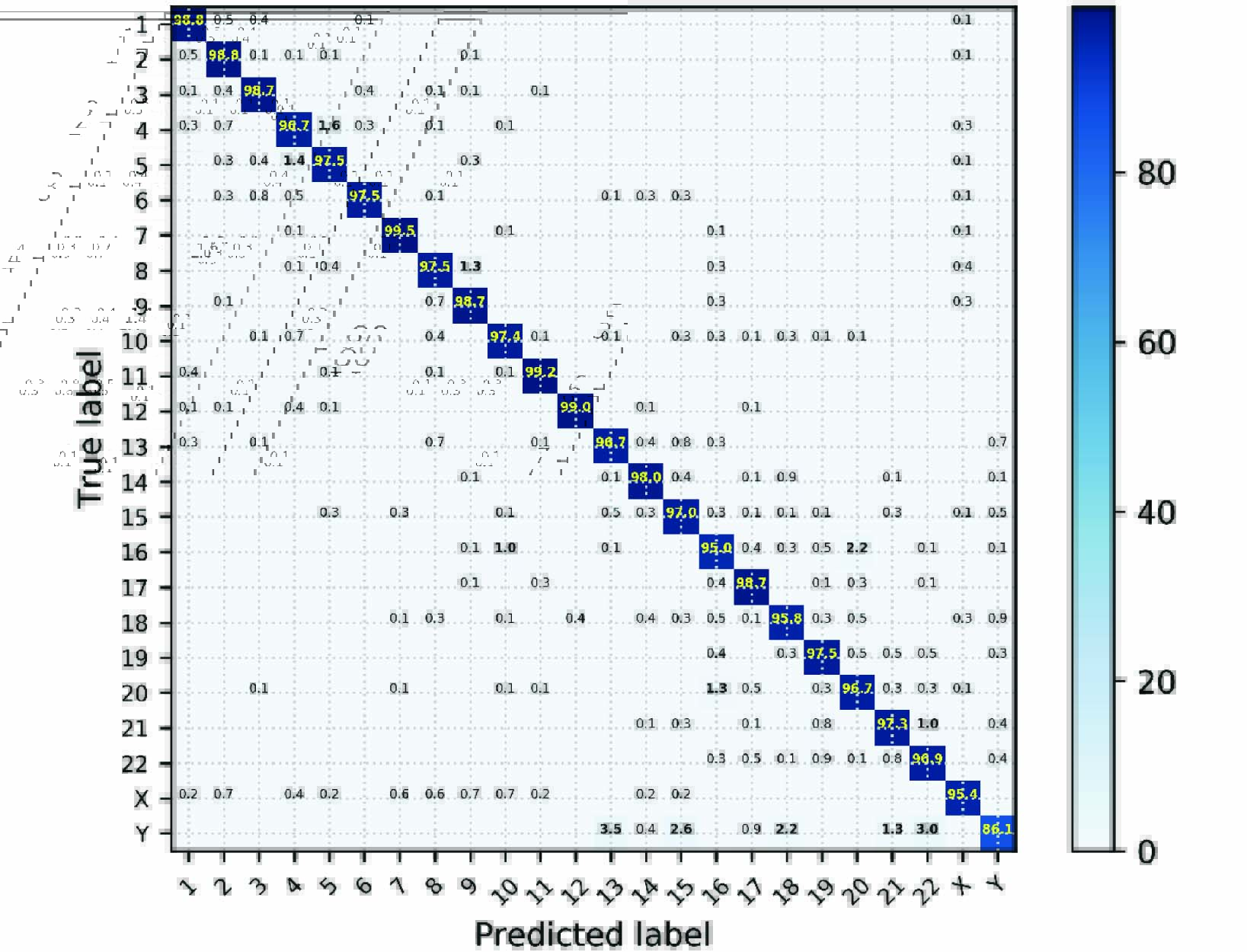}} 
\caption{Confusion matrix of the Varifocal-Net for type classification. The entry in the $i$-th row and $j$-th column denotes the percentage (\%) of the testing samples from class $i$ that were classified as class $j$. Best viewed magnified.}
\label{fig:comfusionmatrix}
\end{figure}

In this section, we present further experiment results of performance analysis. We computed the confusion matrix to get explanatory insights into the results of type prediction. As shown in Fig. \ref{fig:comfusionmatrix}, the confusion between class Y and classes No. 13, No. 15, No. 18, No. 21, and No. 22 mainly contributes to the performance drop.

\begin{figure}[htbp]
\centerline{\includegraphics[width=\columnwidth]{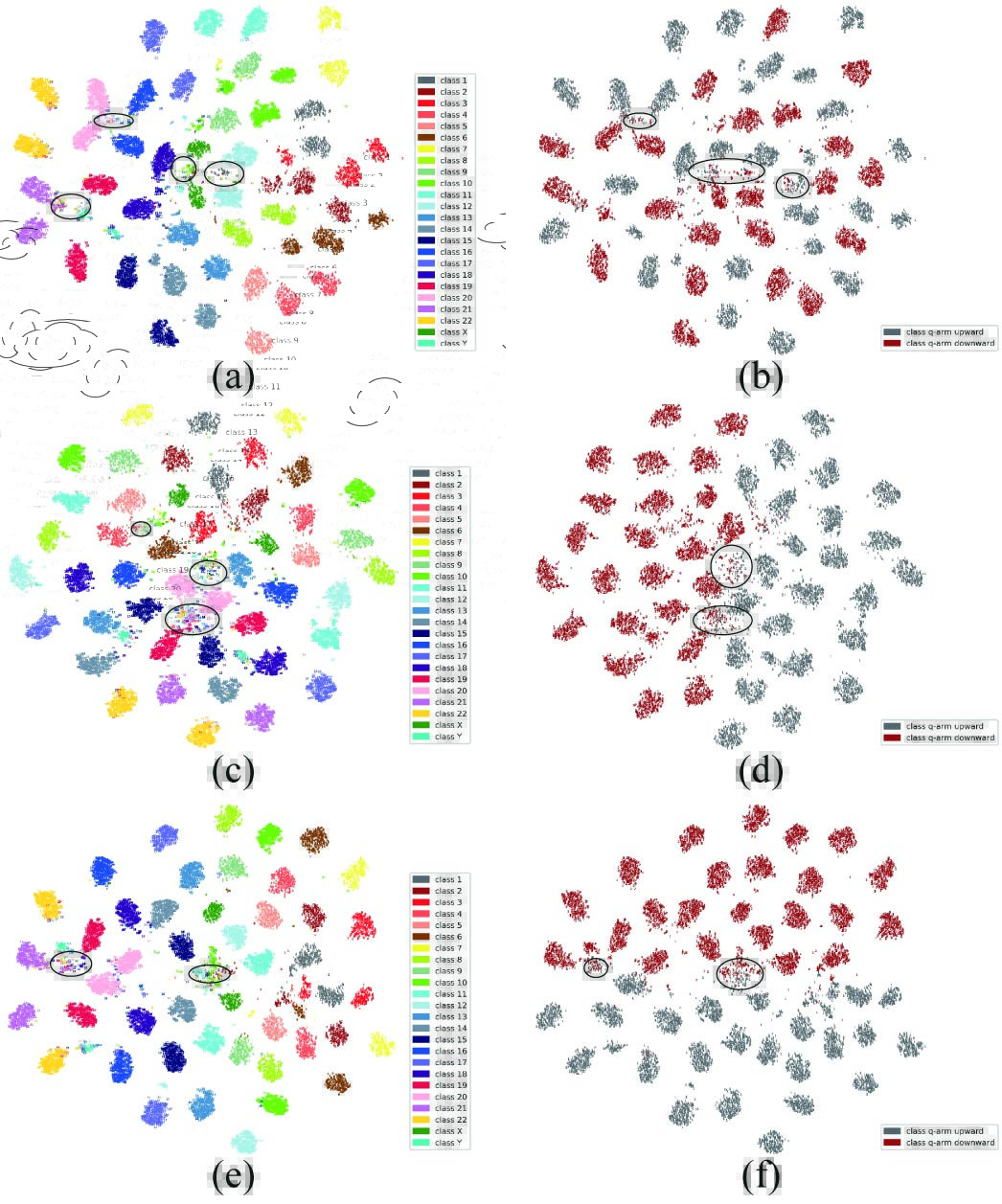}}
\caption{Feature embedding for chromosomes with t-SNE toolbox \cite{maaten2008visualizing}. From the perspective of type classification, the global, local, and concatenated features are visualized in (a), (c), and (e), respectively. Similarly, these three features are visualized in (b), (d), and (f) correspondingly for polarity classification. {The mixed regions of interest are marked with black circles.} Best viewed in color.}
\label{fig:tsne}
\end{figure}

We probed the embedded representations, including the global, local, and concatenated features, in order to illustrate their discrimination capability. We applied the t-SNE \cite{maaten2008visualizing} approach on testing samples' features to reduce their dimensionality {for 2-D visualization. As shown in Fig. \ref{fig:tsne}, the testing samples were clustered by categories and separately dispersed for the concatenated features, with only a small set of samples mixed together. In contrast, for the single-scale global or local features, there exist many large regions where samples of different classes blend together. Compared to Fig. \ref{fig:tsne}(e) and (f), the distance between adjacent clusters in Fig. \ref{fig:tsne}(a)--(d) is smaller. The clusters of global-scale or local-scale features are less compact than that of multi-scale features, making it hard to find a clear boundary for differentiation.}

\begin{figure}[htbp]
\centerline{\includegraphics[width=\columnwidth]{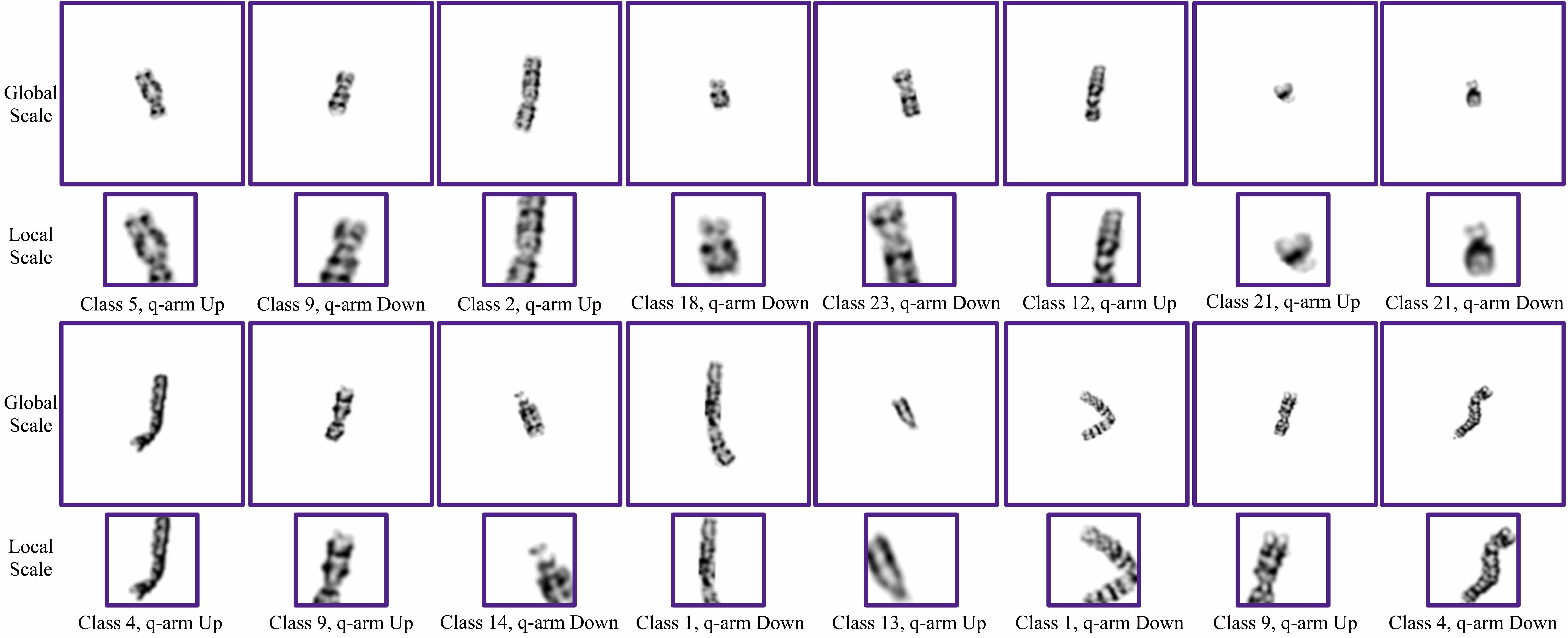}}
\caption{Examples of correctly classified samples. Both global-scale and local-scale inputs are displayed to visually assess the varifocal mechanism.}
\label{fig:egtrue}
\end{figure}

\begin{figure}[htbp]
\centerline{\includegraphics[width=\columnwidth]{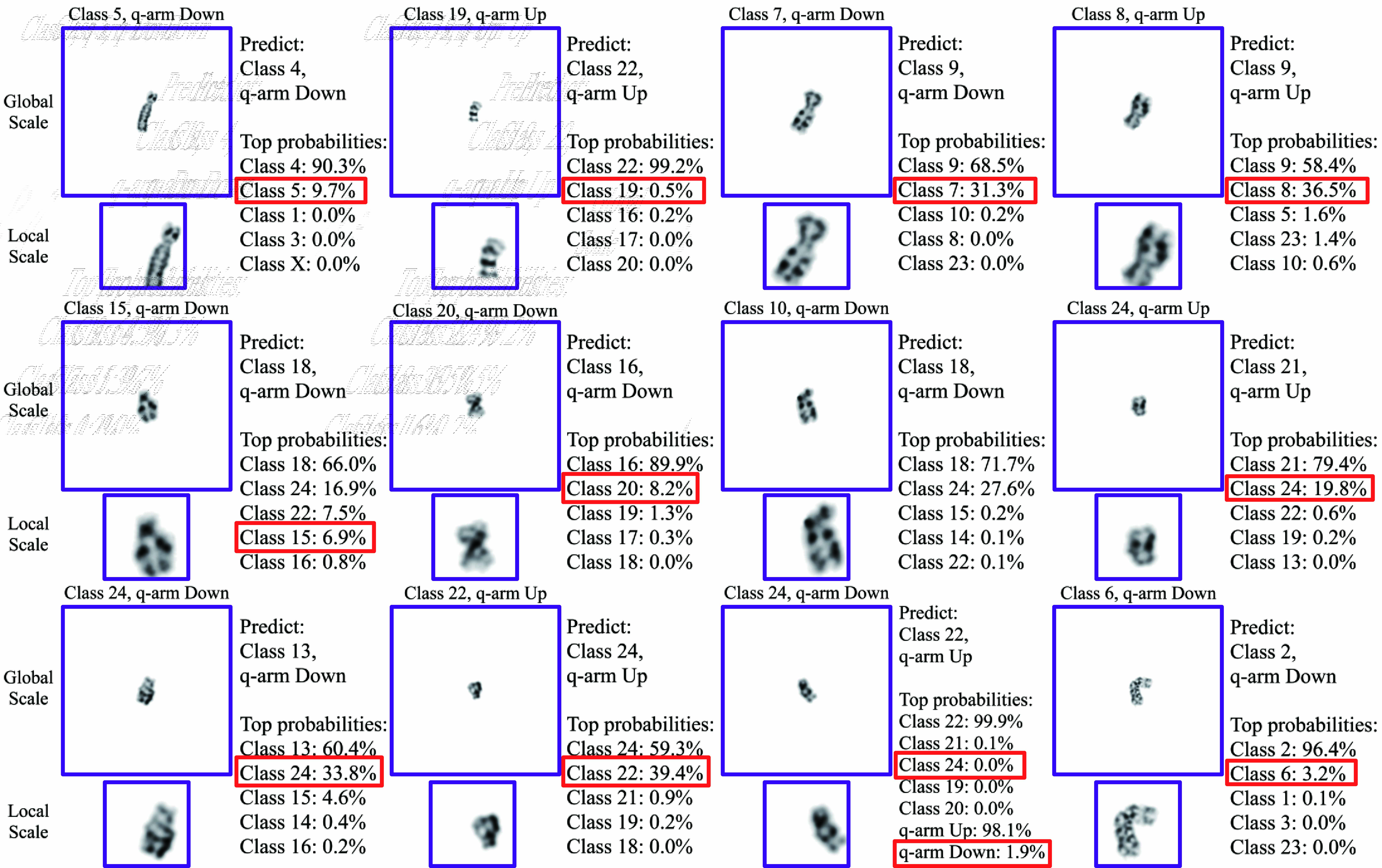}}
\caption{Examples of misclassified samples. The probabilities of wrong predictions are displayed on the right of each image and each red rectangle encloses the predicted probability of the ground-truth label.}
\label{fig:egfalse}
\end{figure}

Figs. \ref{fig:egtrue} and \ref{fig:egfalse} {illustrate} typical examples of correctly and incorrectly classified chromosomes, respectively. Fig. \ref{fig:egtrue} shows that our varifocal mechanism can precisely locate the target region and capture the most discriminative local part with appropriate position and size. For small chromosomes, the predicted box can cover the whole body, while for larger chromosomes, the localization subnet selects partial segments of interest to facilitate accurate recognition. In Fig. \ref{fig:egfalse}, misclassified samples are accompanied with their top 5 probabilities for wrong type predictions and 2 probabilities for wrong polarity predictions. It is observed that for most incorrect predictions, the probability of the true label ranks just the second highest in order. Besides, some chromosomes are grossly distorted or have unusual shapes of their kinds, increasing the difficulty of accurate classification.

\section{Discussion}
\label{sec:discussion}
In this paper, a {three-stage} CNN method was proposed for chromosome classification. Its most distinctive characteristics include: 1) the adoption of varifocal mechanism to detect local discriminative regions; 2) the introduction of residual learning and multi-task learning to facilitate feature extraction; 3) the ensemble of global and local features to boost performance; {4) the use of a dispatch strategy for type assignment in practical karyotyping per case.}

There are mainly two reasons contributing to the inferior performance of the previous CNN-based methods \cite{sharma2017crowdsourcing, gupta2017siamese}. One is the loss of fidelity caused by the straightening step in their pipelines. Although this step is designed to rectify the shape of chromosomes for normalization, it damages the chromosome's morphological consistency and structural information due to inaccurate medial axis extraction and pixel interpolation. In contrast, the proposed Varifocal-Net is an end-to-end method without any shape correction in advance. The other is the lack of large labeled dataset. Their CNNs, which are designed on small datasets, cannot effectively describe the diversity and variety of chromosomes. Hence, these methods lack generality when evaluated on a large testing set.

{As observed from the comparison results in Table \ref{table:comp2}, the potent CNN models \cite{krizhevsky2012imagenet, szegedy2016rethinking, simonyan2014very, he2016deep, huang2017densely, jaderberg2015spatial}} performed well because we adapted them into the same settings as ours. In experiments, we adopted multi-task learning and applied necessary normalization on images. Hence, the performance difference among these models, to a certain extent, reflects the difference of their capabilities of global feature extraction. With respect to their accuracy, there exists a bottleneck of improvement for such single-scale models, which inspired us to resort to multi-scale feature ensemble. With the design of the proposed Varifocal-Net, we keep two aims in mind: the excellent feature extraction ability for classification and the strong discrimination of finer regions detected by the localization subnet. {Since residual units are employed as the backbone of feature extraction CNNs, the proposed method benefits from the introduction of residual learning. Besides, the multi-task learning strategy also contributes to training the network.} For the localization, the varifocal mechanism autonomously focuses on the local part which boosts local feature learning. We can see from Fig. \ref{fig:tsne} that the integration of both global and local features {makes samples of the same category gather closely. It increases between-class distance and reduces chaotic outliers, which explains why our method is superior to the models that only count upon global-scale features.}

Additional comparison on unhealthy cases (see Table \ref{table:comp3}) demonstrated the superior robustness of our method on abnormal chromosome classification. Compared with those single-scale models, our Varifocal-Net utilizes local-scale detail depiction to make up the deficiency of mere consideration of coarse-grained features. {Compared to the Varifocal-Net, there does exist a larger performance decrease for the only G-Net and the only L-Net, which confirms the importance of multi-scale feature ensemble strategy.} Hence, the proposed method, which possesses excellent generalization abilities, can assist doctors in the clinical karyotyping process where abnormal cases occur from time to time.

{The performance improvement of Acc. per Case-D with respect to Acc. per Case in type classification substantiates that the proposed dispatch strategy is effective and suitable for karyotyping within each case. For each method in Table \ref{table:comp2} and Table \ref{table:comp3}, the adoption of the dispatch strategy improves the average accuracy and diminishes the standard deviation for both healthy and unhealthy cases. The generalizability of such strategy lies in the consideration of both maximum likelihood criterion and chromosomal numerical abnormalities.}

The proposed method performed less well on chromosomes No. 15, No. 21, and No. 22 (see Table \ref{table:alltype} and Fig. \ref{fig:comfusionmatrix}) than other classes. Such {three} kinds of chromosomes are acrocentric and contain a segment called satellite, which is separated from the main body. The shape, size, and orientation of satellites differ from one person to another, thus making it difficult for our model to handle all possible situations. Fig. \ref{fig:comfusionmatrix} also shows that chromosome Y is often confused with No. 21 and No. 22. It is because the size and texture of class Y are similar to that of No. 21 and No. 22. Furthermore, the comparatively imbalanced Y samples are not processed with additional data augmentation method, which triggers off poorer recognition of Y. It is noted that although we collected a much larger dataset than previous work, the dataset is still insufficient to cover all possible shapes and sizes of chromosomes. Samples of sex chromosome Y and diversified satellite chromosomes are still in shortage. {Therefore} for better performance, more data should be collected and generative adversarial networks could be used for sample synthesis in the future.

From the results of Table \ref{table:upclass} and examples in Fig. \ref{fig:egfalse}, it is observed that some long chromosomes (e.g., {No. 2 and No. 5}) may be misclassified because their long arms tend to bend or distort greatly during the sampling process. Since the proposed method cannot accurately recognize greatly bent chromosomes, future work may involve particular strategies to cope with this situation. Instead of straightening the chromosomes, we might inform the network of the degree of bending deformation by detecting the rotation pivot (e.g., the centromere) and its angle between two arms. Furthermore, for the G-Net and the L-Net, current feature extractor employs the residual block as a backbone. To further improve performance, we may meticulously redesign the network architecture.

\section{Conclusion}
\label{sec:conclusion}
We have proposed the Varifocal-Net for chromosome classification, which has been evaluated on a large manually constructed dataset. It is a {three}-stage CNN-based method. The first stage effectively learns global and local features through the G-Net and the L-Net, respectively. Taking a global-scale chromosome image as the input, it precisely detects a local region that is discriminative and abundant in details for further feature extraction. The second stage robustly differentiates chromosomes into various types and polarities via two MLP classifiers. It benefits from multi-scale feature ensemble, with only a few misclassifications. {In the third stage, a dispatch strategy was employed to assign each chromosome to a type based on its predicted probabilities.} Extensive experimental results demonstrate that our approach outperforms state-of-the-art methods, corroborating its high accuracy and generalizability.

{Concerning its role in clinical karyotyping workflow, the Varifocal-Net can accurately perform classification within 1 second after operators manually segment chromosomes of a cell for each patient. The karyotyping result maps it automatically generates offer the possibility for human experts to further check and correct possible misclassifications. Moreover, warnings about possible numerical abnormalities allow operators to pay extra attention to the subsequent diagnosis.} {The practical use of the Varifocal-Net in the Xiangya Hospital of Central South University suggests its promising potential for alleviating doctors' workload in the diagnosis process.}

\section*{Acknowledgement}
{The authors are grateful to the anonymous reviewers for their helpful comments.}

\bibliographystyle{IEEEtran}


\end{document}